\documentclass[lettersize,journal]{IEEEtran}
\usepackage{amsmath,amsfonts}
\usepackage{algorithm}
\usepackage{array}

\usepackage[caption=false]{subfig}
\usepackage{textcomp}
\usepackage{stfloats}
\usepackage{url}
\usepackage{verbatim}
\usepackage{graphicx}
\usepackage{cite}
\usepackage{algpseudocode}
\usepackage{hyperref}
\usepackage[dvipsnames]{xcolor}

\usepackage{graphicx}
\usepackage{amsmath}

\usepackage{amssymb}
\usepackage{wasysym}
\usepackage{booktabs}

\usepackage{bm}
\usepackage{multirow}
\usepackage{wrapfig}
\usepackage{listings}
\usepackage[T1]{fontenc}
\usepackage{xcolor}
\usepackage{colortbl}

\newlength\savewidth

\usepackage{array}
\renewcommand{\paragraph}[1]{\noindent\textbf{#1}}

\newcommand{\tablestyle}[2]{\setlength{\tabcolsep}{#1}\renewcommand{\arraystretch}{#2}\centering\footnotesize}

\definecolor{baselinegray}{gray}{.6}
\definecolor{defaultcolor}{gray}{.9}
\newcommand{\default}[1]{\cellcolor{defaultcolor}{#1}}
\definecolor{ourscolor}{HTML}{E0F5E4}

\definecolor{deemph}{gray}{0.6}
\newcommand{\gc}[1]{\textcolor{deemph}{#1}}
\definecolor{decoded_color}{HTML}{A5DCE0}
\definecolor{masked_color}{HTML}{A7A9AC}
\definecolor{expected}{HTML}{3608ED}
\definecolor{uncertain}{HTML}{FF0000}

\definecolor{ratio}{rgb}{0.0, 0.0, 0.0}
\definecolor{sampt}{rgb}{0.0, 0.0, 0.0}
\definecolor{remaskt}{rgb}{0.0, 0.0, 0.0}
\definecolor{cfg}{rgb}{0.0, 0.0, 0.0}
\usepackage{makecell}

\usepackage{pifont}

\usepackage{lipsum}

\newcommand{\pub}[1]{\xspace\textcolor{gray}{\tiny{(\textit{#1})}}}
\definecolor{Highlight}{HTML}{39b54a}  \newcommand{\hl}[1]{\textcolor{Highlight}{\textbf{#1}}}
\newcommand{\dt}[2]{\textbf{#1}\fontsize{5.5pt}{0.1em}\selectfont~\hl{(#2)}}

\usepackage{xspace}
\newcommand{\VQVAE}{VQVAE-2~\cite{razavi2019generating}\pub{NeurIPS'19}\xspace}

\newcommand{\LDM}{LDM~\cite{rombach2022high}\pub{CVPR'22}\xspace}
\newcommand{\UVIT}{U-ViT-H~\cite{bao2022all}\pub{CVPR'23}\xspace}
\newcommand{\DIT}{DiT-XL~\cite{peebles2023scalable}\pub{ICCV'23}\xspace}
\newcommand{\SIT}{SiT-XL~\cite{Ma2024SiTEF}\pub{ECCV'24}\xspace}

\newcommand{\SIMPLEDIFFUSION}{Simple-Diffusion~\cite{hoogeboom2023simple}\pub{ICML'23}\xspace}

\newcommand{\VDMPP}{VDM++~\cite{kingma2023understanding}\pub{NeurIPS'23}\xspace}

\newcommand{\RIN}{RIN~\cite{jabri2022scalable}\pub{ICML'23}\xspace}

\newcommand{\MDT}{MDT~\cite{gao2023masked}\pub{ICCV'23}\xspace}
\newcommand{\LLAMAGEN}{LlamaGen~\cite{sun2024autoregressive}\pub{arxiv'24}\xspace}
\newcommand{\RCGL}{RCG-L~\cite{li2024return}\pub{NeurIPS'24}\xspace}

\newcommand{\VITVQGAN}{ViT-VQGAN~\cite{yu2021vector}\pub{ICLR'22}\xspace}

\newcommand{\VQGAN}{VQGAN~\cite{esser2021taming}\pub{CVPR'21}\xspace}
\newcommand{\VQDiffusion}{VQ-Diffusion~\cite{gu2022vector}\pub{CVPR'22}\xspace}

\newcommand{\TokenCritic}{Token-Critic~\cite{lezama2022improved}\pub{ECCV'22}\xspace}
\newcommand{\DraftRevise}{Draft-and-revise~\cite{lee2022draft}\pub{NeurIPS'22}\xspace}
\newcommand{\MAGE}{MAGE~\cite{li2023mage}\pub{CVPR'23}\xspace}
\newcommand{\MaskgitFSQ}{MaskGIT-FSQ~\cite{mentzer2023finite}\pub{ICLR'24}\xspace}
\newcommand{\Ours}{AdaGen\xspace}

\newcommand{\USF}{USF~\cite{liu2023unified}\pub{ICLR'24}\xspace}

\newcolumntype{x}[1]{>{\centering\arraybackslash}p{#1pt}}
\newcolumntype{y}[1]{>{\raggedright\arraybackslash}p{#1pt}}
\newcolumntype{z}[1]{>{\raggedleft\arraybackslash}p{#1pt}}

\newcommand{\learnable}{\textcolor{learnable}{\textsf{\textit{{learnable}}}}\xspace}
\newcommand{\adaptive}{\textcolor{adaptive}{\textsf{\textit{{adaptive}}}}\xspace}
\definecolor{existing}{HTML}{7F7F7F} \definecolor{learnable}{HTML}{C00000} \definecolor{adaptive}{HTML}{0070C0} \definecolor{imagereward}{HTML}{C00000} \definecolor{agent}{HTML}{1C5CB0} \definecolor{agentcolor}{HTML}{0070BF} \definecolor{rmcolor}{HTML}{EB8253} \definecolor{frozen}{HTML}{6F8EB6}
\definecolor{generatorcolor}{HTML}{F47C7C}

\usepackage{empheq}

\newcommand{\cmark}{\ding{51}}\newcommand{\xmark}{\ding{55}}\usepackage{enumitem}

\usepackage{cases}
\usepackage{xspace}
\makeatletter
\DeclareRobustCommand\onedot{\futurelet\@let@token\@onedot}
\def\@onedot{\ifx\@let@token.\else.\null\fi\xspace}

\def\eg{\emph{e.g}\onedot}

\def\ie{\emph{i.e}\onedot}

\def\etc{\emph{etc}\onedot}
\def\vs{\emph{vs}\onedot}

\makeatother

\usepackage{tikz,xcolor}
\usetikzlibrary{calc, positioning, arrows}
\definecolor{lime}{HTML}{A6CE39}
\DeclareRobustCommand{\orcidicon}{
	\begin{tikzpicture}
		\draw[lime, fill=lime] (0,0)
		circle[radius=0.16]
		node[white]{{\fontfamily{qag}\selectfont \tiny \.{I}D}};
	\end{tikzpicture}
	\hspace{-2mm}
}
\foreach \x in {A, ..., Z}{%
	\expandafter\xdef\csname orcid\x\endcsname{\noexpand\href{https://orcid.org/\csname orcidauthor\x\endcsname}{\noexpand\orcidicon}}
}

\begin{document}

\title{\Ours: Learning Adaptive Policy for \\ Image Synthesis}

\author{Zanlin Ni$^*$, Yulin Wang$^*$, Yeguo Hua, Renping Zhou, Jiayi Guo, Jun Song, Bo Zheng, Gao Huang$^\dagger$
\thanks{Z. Ni, Y. Wang, Y. Hua, R. Zhou, J. Guo, and G. Huang are with the Department of Automation, BNRist, Tsinghua University, Beijing, China. J. Song and B. Zheng are with Taobao \& Tmall Group of Alibaba.}
\thanks{Email: nzl22@mails.tsinghua.edu.cn; gaohuang@tsinghua.edu.cn}
\thanks{$^*$. Equal contribution. $^\dagger$. Corresponding author.}
}

\maketitle

\begin{abstract}
Recent advances in image synthesis have been propelled by powerful generative models, such as Masked Generative Transformers (MaskGIT), autoregressive models, diffusion models, and rectified flow models.
A common principle behind their success is the decomposition of complex synthesis tasks into multiple tractable steps.
However, this introduces a proliferation of step-specific parameters to be configured for modulating the iterative generation process (\eg, mask ratio, noise level, or temperature at each step).
Existing approaches typically rely on manually-designed scheduling rules to manage this complexity, demanding expert knowledge and extensive trial-and-error. Furthermore, these static schedules lack the flexibility to adapt to the unique characteristics of each individual sample, yielding sub-optimal performance.
To address this issue, we present \Ours, a \emph{general}, \emph{learnable}, and \emph{sample-adaptive} framework for scheduling the iterative generation process. 
Specifically, we formulate the scheduling problem as a Markov Decision Process, where a lightweight policy network is introduced to adaptively determine the most suitable parameters given the current generation state, and can be trained through reinforcement learning.
Importantly, we demonstrate that simple reward designs, such as FID or pre-trained reward models, can be easily hacked and may not reliably guarantee the desired quality or diversity of generated samples.
Therefore, we propose an adversarial reward design to guide the training of the policy networks effectively.
Finally, we introduce an inference-time refinement strategy and a controllable fidelity-diversity trade-off mechanism to further enhance the performance and flexibility of \Ours.
Comprehensive experiments across five benchmark datasets (ImageNet-256$\times$256 \& 512$\times$512, MS-COCO, CC3M, and LAION-5B) and four distinct generative paradigms validate the superiority of \Ours{}.
For example, \Ours achieves better performance on DiT-XL with $\mathbf{\sim 3\times}$ lower inference cost and improves the FID of VAR from 1.92 to {1.59} with negligible additional computational overhead.
Code and pre-trained models are available at \url{https://github.com/LeapLabTHU/AdaGen}.
\end{abstract}

\begin{IEEEkeywords}
Adaptive image generation, reinforcement learning, iterative generation models
\end{IEEEkeywords}

\section{Introduction}
\label{sec:introduction}

\begin{figure}[t!]\centering
	\includegraphics[width=\linewidth]{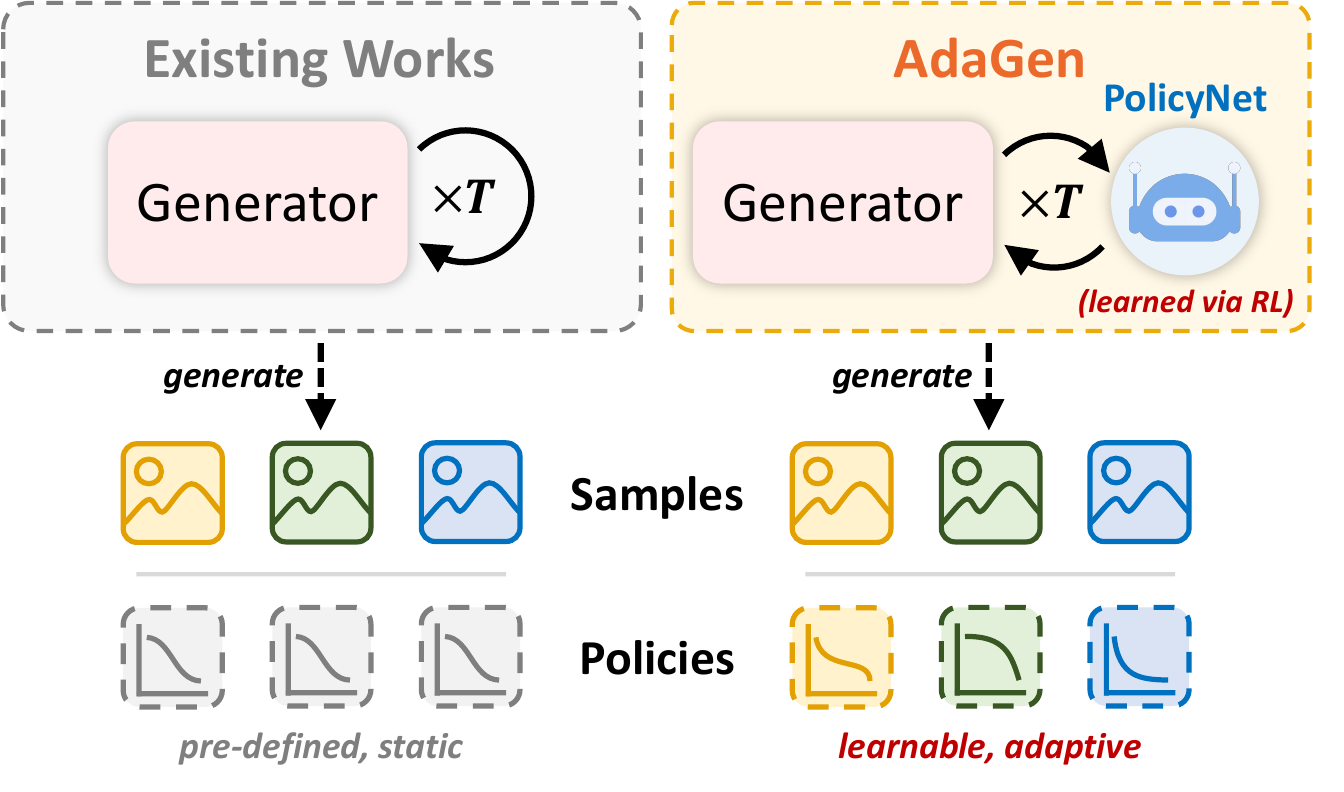}
	\vskip -.15in
	\caption{
\textbf{Main idea of \Ours.} 
Existing multi-step generative models, such as diffusion models, typically utilize \emph{pre-defined, static} schedules to configure the generation policy (\eg, noise level) of all samples. 
Instead, \Ours leverages reinforcement learning (RL) to train a policy network that directly \emph{learns} the optimal generation policies \emph{adaptively} tailored for each sample.
}	
	\label{fig:method_motivation}
	\vspace{-0.9em}
\end{figure}

\begin{table*}[t]
	\centering
	\caption{Summary of Generation Policies for Different Generation Paradigms
	}
	\vspace{-.7em}
	\label{tab:hyperparam_summary}
	\tablestyle{1pt}{1.1}
	\begin{tabular}{y{70}y{100}y{80}x{10}y{220}}
		\toprule
		Type & Generation Policy & Range & \multicolumn{2}{l}{Scheduling Rules} \\
		\midrule
		\multirow{4}{*}{MaskGIT} 
		& Mask ratio, \(m_t\)  & \(m_t \in [0,1]\)   & \(m_t\) & \(= \cos(0.5 \pi \cdot t / T)\)~\cite{chang2022maskgit,chang2023muse}, \(2\arccos(t/T)/\pi\)~\cite{besnier2023pytorch,yu2024image}\\
		& Sampling temperature, \(\tau_t\)  & \(\tau_t \ge 0\)  & \(\tau_t\) & \(= 1\)~\cite{chang2022maskgit,chang2023muse}, \(0.5 + 0.8 \times (1 - t/T)\)~\cite{yu2024image}\\
		& Masking temperature, \(\zeta_t\)  & \(\zeta_t \ge 0\)  & \(\zeta_t\) & \(= C \cdot (1-t/T)\)~\cite{chang2022maskgit,chang2023muse}\\
		& Guidance scale, \(w_t\)  & \(w_t \ge 0\)  & \(w_t\) & \(= C\cdot t / T\)~\cite{chang2022maskgit,chang2023muse}, \(C(1 - \cos \pi \left( {t}/{T} \right)^{C'}) / 2 \)~\cite{yu2024image} \\
		\midrule
		\multirow{4}{*}{Autoregressive} 
		& Sampling temperature, \(\tau_t\)  
		& \(\tau_t \ge 0\)  
		& \(\tau_t\) & \(= C\)~\cite{Tian2024VisualAM,sun2024autoregressive} \\
		& Guidance scale, \(w_t\)  
		& \(w_t \ge 0\)  
		& \(w_t\) & \(=C\)~\cite{sun2024autoregressive}, \(C\cdot t / T\)~\cite{Tian2024VisualAM} \\
		& Top-\(k\) parameter, \(k_t\)  
		& \(k_t \in \mathbb{N}^+\)  
		& \(k_t\) & \(= C\)~\cite{Tian2024VisualAM,sun2024autoregressive} \\
		& Top-\(p\) parameter, \(\rho_t\)  
		& \(\rho_t \in [0,1]\)  
		& \(\rho_t\) & \(= C\)~\cite{Tian2024VisualAM,sun2024autoregressive} \\
		\midrule
		\multirow{2}{*}{Diffusion (ODE)} 
		& Timestep, \(\kappa_t\)  
		& \(\kappa_t \in \{0,\dots,\kappa_{\max}\}\)  
		& \(\kappa_t\) & \(=\lfloor (1-t/T)\cdot \kappa_{\max}\rfloor\)\cite{song2020denoising,lu2022dpmp},\ \(\ \lfloor (1-t/T)^2\cdot \kappa_{\max}\rfloor\)\cite{song2020denoising,lu2022dpmp} \\
		& Guidance scale, \(w_t\)  
		& \(w_t \ge 0\)  
		& \(w_t\) & \(= C\) \cite{dhariwal2021diffusion,bao2022all,peebles2023scalable}, \(C(1 - \cos \pi \left( {\kappa_t}/{\kappa_{\max}} \right)^{C'}) / 2 \)~\cite{gao2023masked} \\
		\midrule
		\multirow{2}{*}{Rectified Flow} 
		& Timestep, \(\kappa_t\)  
		& \(\kappa_t \in [0,1]\)  
		& \(\kappa_t\) & \(= 1-t/T\)~\cite{Ma2024SiTEF} \\
		& Guidance scale, \(w_t\)  
		& \(w_t \ge 0\)  
		& \(w_t\) & \(= C\)~\cite{Ma2024SiTEF} \\
		\bottomrule
	\end{tabular}
	\begin{minipage}{.94\linewidth}
		\vspace{.4em}
		\scriptsize
		For each paradigm, we list the policy names, their value ranges, and the common settings adopted.
		We use $t\in\{1,\dots,T\}$ as a unified index representing the current generation step.
		For diffusion models, we focus on ODE-based diffusion sampling (denoted as Diffusion (ODE)) following recent works~\cite{song2020denoising,song2021maximum,lu2022dpm,lu2022dpmp}.
		We note that one should distinguish the \emph{generation step $t$} from the policy parameter \emph{timestep $\kappa_t$}. The latter arises in specific paradigms such as diffusion and rectified flow models, where it serves as a discretization point for numerical ODE solvers and implicitly controls the noise level at the $t$-th generation step~\cite{lu2022dpm}.
		This timestep parameter is conventionally denoted by $t_i$~\cite{lu2022dpm,lu2022dpmp,liu2023unified}. To avoid confusion with generation step $t$, we use $\kappa_t$ instead.
		$C, C'$ are constants.
	\end{minipage}
	\vspace{-0.8em}
\end{table*}

Over the past few years, image synthesis has experienced dramatic advancements through the emergence of powerful generative models.
Language-inspired frameworks such as Masked Generative Transformers (MaskGIT)~\cite{chang2022maskgit,lezama2022improved,chang2023muse,li2023mage} and autoregressive models~\cite{esser2021taming,yu2022scaling,Tian2024VisualAM,sun2024autoregressive} conceptualize image generation as a form of token modeling, either by predicting masked tokens or by sequentially sampling one token at a time. 
Concurrently, generative paradigms like diffusion models~\cite{dhariwal2021diffusion,rombach2022high,peebles2023scalable} and rectified flow methods~\cite{liu2022flow,Ma2024SiTEF} formulate generation as an iterative denoising process guided by learned score or velocity functions. 
These diverse modeling paradigms have all demonstrated impressive scalability and the ability to produce high-fidelity visual content.

Despite different underlying mechanisms, these methodologies share a key principle: decomposing the complex task of photorealistic image generation into a sequence of more manageable sub-problems.
For instance, MaskGIT only decodes a subset of masked tokens per step, while diffusion models iteratively remove small amounts of noise from the sample.
While such multi-step strategies have proven instrumental in their success, they also create a significant challenge: a proliferation of step-conditional parameters, which we collectively term the \emph{generation policy}, must be properly configured to ensure their effectiveness.
For example, at each step, diffusion models need to determine an appropriate noise reduction level, while MaskGIT requires proper calibration of masked token proportions.
As the number of steps increases, the configuration complexity may quickly become unmanageable.
Existing approaches tackle this challenge by pre-defining multiple scheduling functions (see Table~\ref{tab:hyperparam_summary}), which require expert knowledge and may still fail to yield optimal results.
Moreover, a globally shared policy may lack the flexibility to adapt to the unique characteristics of individual samples.

To address these limitations, this paper presents \Ours. As illustrated in Figure~\ref{fig:method_motivation}, our key insight is to consider a \emph{learnable} policy network to automatically configure the proper generation policy \emph{adaptively} tailored for each sample. 
By treating policy design as a data-driven optimization problem rather than a manual art, our approach removes the burden of expert scheduling and enables per-sample customization.
However, training this policy network is challenging as end-to-end backpropagation through the entire multi-step generation process is computationally infeasible.
To overcome this, we formulate the determination of the optimal generation policy as a Markov decision process (MDP). 
Within this formulation, the policy network naturally becomes an agent that observes the current generation state, adaptively decides the policy to maximize final quality, and can be trained through reinforcement learning (\emph{e.g.}, policy gradient).

Another challenge in our problem lies in designing effective reward signals.
We find that straightforward approaches, such as employing standard metrics (\eg, FID) or pre-trained reward models~\cite{xu2024imagereward}, often lead to unintended outcomes.
More specifically, although the expected value of rewards can be successfully maximized, the resulting policies usually guide the generation process to produce insufficiently high-fidelity and diverse images (see Figure~\ref{fig:comp_rewards}).
In other words, the policy network tends to ``overfit'' these rewards.
Inspired by this phenomenon, we hypothesize that a potential solution is to consider a reward that is \emph{updated concurrently with the policy network} and \emph{adjusted dynamically to resist overfitting}. 
Therefore, we propose an adversarial reward model, which is formulated as a discriminator similar to that used in generative adversarial networks (GANs)~\cite{goodfellow2014generative}.
When the policy network learns to maximize the reward, we refine the reward model simultaneously to better distinguish between real and generated samples.
In this way, the policy network is effectively prevented from overfitting a static objective, and we observe a balanced diversity and fidelity of the generated images.
The training pipeline of \Ours is illustrated in Figure~\ref{fig:pipeline}.

Beyond the core framework, we additionally introduce two enhancements to \Ours.
\emph{First}, we propose an inference-time refinement strategy for further generation quality improvement.
Specifically, we find that the auxiliary networks in \Ours (\eg, the adversarial reward model), though originally discarded after training, can be repurposed as off-the-shelf perceptual evaluators.
Leveraging their predictions, we design refinement mechanisms that guide sampling toward higher-quality outputs, without requiring additional training or external models (Table~\ref{tab:inference_refinement}, Table~\ref{tab:inference_refinement2}).
\emph{Second}, we enable controllable trade-offs between fidelity and diversity in \Ours.
Specifically, we introduce a fidelity-oriented policy network and blend its output with the original policy via a scalar $\lambda$.
In the meantime, we blend our adversarial reward and a fidelity-centric reward using the same $\lambda$ for the reward signal.
This dual-interpolation design explicitly bridges the control parameter $\lambda$ with the fidelity-diversity spectrum, thereby offering a user-controllable modulation of generation behavior.

The effectiveness of \Ours{} is comprehensively validated across five benchmark datasets: ImageNet 256$\times$256~\cite{russakovsky2015imagenet}, ImageNet 512$\times$512~\cite{russakovsky2015imagenet}, MS-COCO~\cite{lin2014microsoft}, CC3M~\cite{sharma2018conceptual}, and LAION-5B~\cite{schuhmann2022laion}. We demonstrate the generality of \Ours{} by applying it to five representative generative models spanning four distinct paradigms: MaskGIT~\cite{chang2022maskgit}, DiT~\cite{peebles2023scalable} (diffusion), SiT~\cite{Ma2024SiTEF} (rectified flow), VAR~\cite{Tian2024VisualAM} (autoregressive), and Stable Diffusion~\cite{rombach2022high}.
\Ours{} achieves significant computational efficiency and performance improvements across diverse paradigms, delivering 17\% to 54\% improvements in generation performance or 1.6$\times$ to 3.6$\times$ reductions in inference cost when maintaining comparable generation performance.

This paper is an extended version of our conference paper AdaNAT~\cite{Ni2024AdaNAT}. We make several important new contributions:

\begin{itemize}
	\item We extend the original AdaNAT to a more general and flexible framework, \Ours. \Ours extends the idea of learning adaptive generation policies beyond MaskGIT, encompassing major multi-step generative paradigms such as diffusion, autoregressive, and rectified flow models (Section~\ref{sec:ours}), and achieves competitive performance across them. Meanwhile, we ensure more stable policy optimization as generation steps increase by proposing an action smoothing technique (Section~\ref{sec:action_stablize}).
	\item We introduce an off-the-shelf inference-time refinement strategy for further generation quality enhancement (Section~\ref{sec:inference_refinement}). Specifically, we find that the auxiliary networks in \Ours can be effectively repurposed as perceptual evaluators. Guided by their predictions, we develop a refinement mechanism that enhances generation quality without requiring retraining or external models.
	\item We enable explicit control over the fidelity-diversity trade-off in \Ours (Section~\ref{sec:tradeoff}), by introducing a fidelity-centric policy network and blending its output with our original policy via a user-controllable scalar.
	\item We conduct extensive experiments demonstrating \Ours's efficacy across four generative paradigms and large-scale settings, including 2B-parameter VAR~\cite{Tian2024VisualAM} and Stable Diffusion~\cite{rombach2022high} on LAION-5B~\cite{schuhmann2022laion} (Tables~\ref{tab:fid_add_adagen},~\ref{tab:fid_imnet},~\ref{tab:fid_mscoco_zero_shot}). We also validate the feasibility of our newly proposed inference-time refinement and fidelity-diversity control mechanisms (Tables~\ref{tab:inference_refinement},~\ref{tab:inference_refinement2}, Figure~\ref{fig:tradeoff_fid_div}).
	\item We provide more comprehensive analyses of \Ours covering computational overhead, hyperparameter sensitivity, and extensive ablation studies on our key components (Tables~\ref{tab:computational_overhead},~\ref{tab:ppo_hyp},~\ref{tab:abl_momentum},~\ref{tab:ablations}).
\end{itemize}

\section{Related Work}

\paragraph{Iterative Generation Models}
have now largely replaced traditional Generative Adversarial Networks (GANs)~\cite{goodfellow2014generative,karras2019style,Karras2019stylegan2} as the dominant approach for high-quality image synthesis, due to their superior scalability~\cite{saharia2022photorealistic,yu2022scaling,kang2023scaling,peebles2023scalable} and sample diversity~\cite{dhariwal2021diffusion,saharia2022photorealistic,ramesh2022hierarchical}.
Broadly speaking, these models can be categorized into two main families: discrete token-based and continuous-state approaches.
Token-based models, inspired by advances in language modeling, represent images as sequences of discrete visual tokens and generate them either autoregressively~\cite{esser2021taming,ding2021cogview,yu2022scaling,Tian2024VisualAM}, predicting each token conditioned on previous ones, or through masked modeling strategies~\cite{chang2022maskgit,lezama2022improved,chang2023muse,li2023mage}, where subsets of tokens are iteratively predicted while others are held fixed.
Continuous-state models, such as diffusion models, iteratively refine a noisy input in pixel or latent space by reversing a noise process~\cite{dhariwal2021diffusion,rombach2022high,saharia2022photorealistic,peebles2023scalable}, while rectified flow models use deterministic velocity fields for faster, more stable sampling~\cite{liu2022flow,Ma2024SiTEF}.
Despite their differences, these frameworks share the goal of approximating complex data distributions through incremental refinements. Our work builds on this by introducing a lightweight policy network that adaptively predicts per-step generation policies, improving generative model output quality without tuning the model itself.

\paragraph{Reinforcement Learning in Image Generation.}
The integration of reinforcement learning (RL) for image generation began with early works like~\cite{bachman2015data}, which explored policy-based approaches to structured prediction tasks. Recently, ImageReward~\cite{xu2024imagereward} collected a large-scale human preference dataset to train a reward model, establishing a framework for quantifying image quality based on human aesthetic judgments. This pre-trained reward model has spurred research on using RL to fine-tune diffusion-based image generation models~\cite{black2023training,fan2024reinforcement,zhang2024large}, where the reward signal guides the generative model toward synthesizing images with certain desired properties.
Unlike these methods, which optimize the generative model itself through parameter updates, our work employs an RL agent to enhance a frozen generative model without modifying its internal parameters.
This formulation shifts the focus from generator optimization to generation control, and obviates the need for costly retraining or fine-tuning of the base generator.

\paragraph{Generative Adversarial Networks and RL.}
Several works have used ideas from reinforcement learning to train GANs~\cite{yu2017seqgan,wang2017irgan,wang2018no,sarmad2019rl} for text generation, information retrieval, point cloud completion, \etc.
Recently, SFT-PG~\cite{fan2023optimizing} combines RL with a GAN objective for diffusion-based image generation. Our work differs from these works in many important aspects.
First, the research problem is orthogonal. We are interested in exploring a better generation policy \emph{given} a pre-trained generator backbone (which is kept unchanged throughout our method), while previous studies investigate the alternative ``RL+GAN'' approach to directly train or fine-tune the generator backbone itself.
Second, our method is general and applicable to various iterative generation models, while previous works focus on improving specific generative paradigms, such as diffusion models.
Third, compared to SFT-PG~\cite{fan2023optimizing}, we conduct comparative analyses on different reward designs and large-scale experiments while they mainly explored the GAN-based objective and proved their concept on relatively simple datasets (\eg, CIFAR~\cite{krizhevsky2009learning} and CelebA~\cite{liu2015deep}).

\paragraph{Proper Scheduling for Generative Models.}
Due to the step-dependent nature of iterative models' generation policies, designing proper sampling schedules has emerged as a critical factor for effective generation.
Traditionally, these schedules are set heuristically, relying on expert intuition or extensive trial-and-error. For example, MaskGIT~\cite{chang2022maskgit} compared several mask ratio schedules before settling on a cosine schedule, while MDT~\cite{gao2023masked} crafted a power-cosine schedule for classifier-free guidance. However, such manual approaches are labor-intensive and may fail to yield the most suitable policies.
To address this, recent works have explored ways to automatically obtain these schedules via differentiating through sample quality~\cite{Watson2022LearningFS}, evolutionary search~\cite{li2023autodiffusion}, numerical gradients~\cite{ni2024revisiting}, and predictor-based search~\cite{liu2023oms}.
Despite these advancements, these automatic scheduling methods exhibit certain limitations.
First, they typically employ static objectives (\eg, standard evaluation metrics~\cite{heusel2017gans,binkowski2018demystifying}), a practice which, as we identify in Section~\ref{sec:reward}, can be susceptible to overfitting issues.
Second, these methods are often tailored to specific generative paradigms, predominantly diffusion models, leaving other iterative models underexplored.
Third, the learned generation policies are globally shared, ignoring individual sample characteristics.
In contrast, our proposed adversarial learning paradigm is designed to mitigate reward overfitting. Moreover, our framework is general and applicable to various iterative generation models, and our adaptive policy network furnishes sample-specific policies, enabling generated outputs that exhibit both enhanced quality and greater diversity.

\section{Method}
This section presents the \Ours methodology. We begin by reviewing four generative paradigms and their policies in Section~\ref{sec:prelim}, followed by our motivation in Section~\ref{sec:motivation}. We then introduce our core MDP framework in Section~\ref{sec:ours} and reward design in Section~\ref{sec:reward}. Section~\ref{sec:action_stablize} presents a technique for stabilizing policy network training with larger generation steps, while Section~\ref{sec:improvements} details additional enhancements including inference-time refinement and a blending mechanism for balancing fidelity and diversity. Details of the experiments in this section are available in Appendix~\ref{sec:app_exp_details}.

\subsection{Preliminaries}
\label{sec:prelim}
We briefly review the four mainstream paradigms for generative modeling presented in Table~\ref{tab:hyperparam_summary}. All paradigms aim to model a target data distribution $p(\bm{x})$ through a parametric model $p_{\bm{\theta}}$ with learnable parameters $\bm{\theta}$.

\paragraph{Masked Generative Transformers} (MaskGIT)~\cite{chang2022maskgit,chang2023muse} learn to generate samples using a masked token modeling objective~\cite{devlin2019bert}. Once trained, the model can predict masked tokens based on their surrounding unmasked context, enabling iterative and parallel token generation.
During inference, the process begins with a fully masked token sequence $\bm{x}_0$. At each generation step $t$, the model takes the partially masked sequence $\bm{x}_t$ as input, predicts all masked tokens in parallel, and then selectively re-masks tokens with low confidence: $p_{\bm{\theta}}(\bm{x}_{t+1}|\bm{x}_t) = \sum_{\hat{\bm{x}}_{t+1}} p_{\bm{\theta}}^{\mathrm{pred}}(\hat{\bm{x}}_{t+1}|\bm{x}_t) \cdot p_{\bm{\theta}}^{\mathrm{remask}}(\bm{x}_{t+1}|\hat{\bm{x}}_{t+1})$.
This iterative process continues until all tokens are decoded.

Typically, their generation policy consists of four components~\cite{chang2022maskgit,chang2023muse}: the ratio of tokens to be masked for the next step $m_t$, the temperature for token sampling $\tau_{t}$ and token masking $\zeta_t$, and the classifier-free guidance scale $w_{t}$.

\paragraph{Autoregressive Models}~\cite{sun2024autoregressive,Tian2024VisualAM} generate data through sequential conditional modeling, where each generation step depends on all previously generated content:
$p_{\bm{\theta}}(\bm{x}) = \prod_{t=1}^{T} p_{\bm{\theta}}(\bm{x}_t | \bm{x}_{<t})$.
Here, $\bm{x}_{<t}$ denotes all information generated prior to step $t$, while $\bm{x}_t$ represents the autoregressive data unit generated at step $t$, which is typically a single token.
Recently, VAR~\cite{Tian2024VisualAM} shows that treating ``scale'' as the autoregressive unit and performing next-scale prediction offers a more natural approach for visual generation.

Typically, their generation policy consists of four components~\cite{sun2024autoregressive,Tian2024VisualAM}: the temperature \( \tau_t \) for token sampling, the classifier-free guidance scale \( w_t \), the top-k filtering parameter \( k_t \), and the top-p nucleus sampling parameter \( \rho_t \).

\paragraph{Diffusion Models}~\cite{ho2020denoising,peebles2023scalable} learn to generate data by reversing a gradual noising process. The noise process starts from real data \( \bm{x}_0 \), noise is progressively added to obtain \( \bm{x}_t \)\footnote{Here we slightly overload the notation $t$ to follow the conventional notation in diffusion models. This should be distinguished from the generation step $t$ in our main text, which indexes the sequential generation iterations.} via
\( q(\bm{x}_t | \bm{x}_0) = \mathcal{N}(\bm{x}_t; \sqrt{\bar{\alpha}_t} \bm{x}_0, (1 - \bar{\alpha}_t) \bm{I}) \),
where \( \bar{\alpha}_t \) are pre-defined hyperparameters. The model is trained to approximate the reverse, denoising process:
\( p_{\bm{\theta}}(\bm{x}_{t-1} | \bm{x}_t) = \mathcal{N}(\bm{x}_{t-1}; \mu_{\bm{\theta}}(\bm{x}_t, t), \sigma_t^2 \bm{I}) \),
typically using the following loss function when parameterizing \( \mu_{\bm{\theta}} \) as a noise-prediction network \( \epsilon_{\bm{\theta}} \):
\( \mathcal{L}_{\text{simple}}(\bm{\theta}) = \|\epsilon_{\bm{\theta}}(\bm{x}_t) - \bm{\epsilon}\|_2^2 \).
To generate a sample, the process starts from pure noise \( \bm{x}_{t_{\text{max}}} \sim \mathcal{N}(0, \bm{I}) \) and iteratively samples
\( \bm{x}_{t-1} \sim p_{\bm{\theta}}(\bm{x}_{t-1} | \bm{x}_t) \) to recover data.
Recent work~\cite{song2020denoising,song2021maximum,lu2022dpm,lu2022dpmp} shows that ODE-based sampling for diffusion models greatly speeds up generation with little loss in quality. We follow this approach in our method.

Typically, their generation policy consists of two components~\cite{lu2022dpm,lu2022dpmp}: the timestep \( \kappa_t \) for solving the ODE, and the classifier-free guidance scale \( w_t \), which modifies the predicted noise as $\epsilon_{\bm{\theta}, w_t}(\bm{x}_t) = (1 + w_t) \epsilon_{\bm{\theta}}(\bm{x}_t) - w_t \epsilon_{\bm{\theta}}^{\text{uncond}}(\bm{x}_t)$.

\paragraph{Rectified Flow Models}~\cite{liu2022flow,Ma2024SiTEF} generate samples by progressively transforming a prior distribution, typically a standard normal, into the target data distribution through a continuous flow. The forward process is defined as a linear interpolation between a noise sample \( \bm{x}_1 \sim \mathcal{N}(0, \bm{I}) \) and a real data sample \( \bm{x}_0 \):
\( \bm{x}_t = t\bm{x}_1 + (1 - t)\bm{x}_0, \; t \in [0,1] \).
The model learns a velocity field \( v_{\bm{\theta}} \) that predicts how the sample evolves along this trajectory:
\( \min_{\bm{\theta}} \mathbb{E}_{t, \bm{x}_0, \bm{x}_1} \big\| v_{\bm{\theta}}(\bm{x}_t, t) - (\bm{x}_1 - \bm{x}_0) \big\|_2^2 \).
Starting from \( \bm{x}_1\sim \mathcal{N}(0, \bm{I}) \) (corresponding to \( t = 1 \)), samples are generated by integrating the learned velocity field via an ordinary differential equation:
\( \frac{d\bm{x}_t}{dt} = v_{\bm{\theta}}(\bm{x}_t, t), \; t: 1 \rightarrow 0 \).

Similar to ODE-based diffusion sampling, the generation policy of rectified flow models also consists of two components~\cite{Ma2024SiTEF}:
the timestep \( \kappa_t \) for solving the ODE, and the classifier-free guidance scale \( w_t \) for a modified velocity prediction function $v_{\bm{\theta}, w_t}$.

\subsection{Motivation}
\label{sec:motivation}

\vskip -.2in
 \begin{align}
 	\textcolor{baselinegray}{\textsf{Existing Policy}:} \quad & \textcolor{baselinegray}{\bm{\eta}(t)} & \textcolor{baselinegray}{\textsf{\textit{pre-defined}, \textit{static}}} \label{eq:exist} \\
 	\text{\textbf{\textsf{\Ours Policy}}:} \quad & \bm{\eta}_{\textcolor{learnable}{{\bm{\phi}}}}(\textcolor{adaptive}{\bm{s}_t}) & \text{\learnable, \adaptive} \label{eq:optimal}
 \end{align}

\paragraph{Motivation I: Automatic Policy Acquisition.}
Modern generative paradigms sequentially decomposes the complex generation process into temporally discretized steps, which inherently introduces temporal dependency in their parameterization. Moreover, as discussed in Section~\ref{sec:prelim}, these generative models typically require more than one type of generation policy to control different aspects of the generation dynamics. This dual challenge of temporal dependency and policy diversity combinatorially amplifies configuration complexity. For example, even with a moderate $T=32$ generation steps, MaskGIT would require 128 policy parameters to be configured, making it difficult for practitioners to fully exploit these models in realistic scenarios. Existing approaches mainly alleviate this problem by leveraging multiple \emph{pre-defined} scheduling rules (collectively denoted as $\bm{\eta}(t)$, detailed in Table~\ref{tab:hyperparam_summary}), which can be either manually designed scheduling functions or simply fixed values across all generation steps.
However, manual designs require substantial expert knowledge and effort, while over-simplified strategies may fail to capture the nuanced needs across generation steps.
To address this issue, we propose to consider a learnable policy network $\bm{\eta}_{\bm{\phi}}$ that is trained to produce the appropriate policy automatically.
As a result, a significantly better generation policy can be acquired with minimal human effort once a proper learning algorithm for $\bm{\eta}_{\bm{\phi}}$ is utilized (which will be discussed in Section~\ref{sec:ours}).

\paragraph{Motivation II: Adaptive Policy Adjustment.}
Furthermore, it is noteworthy that Eq.~\eqref{eq:exist} is a \emph{static} formulation, \ie, the generation of all samples shares the same set of scheduling functions.
In contrast, we argue that every sample has its own characteristics, and ideally the generation process should be adaptively adjusted according to each individual sample.
To attain this goal, we propose to dynamically determine the policy for $t^{\textnormal{th}}$ generation step conditioned on the current generation status $\bm{s}_t$ (detailed in Table~\ref{tab:mdp_unified}).
In other words, $\bm{s}_t$ provides necessary information on how the generated sample `looks like' at $t^{\textnormal{th}}$ step, based on which a tailored policy can be derived to enhance the generation quality.
More evidence to support our idea can be found in Table~\ref{tab:effectiveness} and Figure~\ref{fig:vis_policy}.

Driven by the discussions above, we propose to introduce a policy network $\bm{\eta}_{\bm{\phi}}$ that directly \emph{learns} to produce the appropriate policy in an \emph{adaptive} manner, as shown in Eq.~\eqref{eq:optimal}. In the following, we will introduce how to train $\bm{\eta}_{\bm{\phi}}$ effectively.

\subsection{Unified MDP Formulation}
\label{sec:ours}
Formally, given a pre-trained generative model parameterized by $\bm{\theta}$ and a policy network $\bm{\eta}_{\bm{\phi}}$ to be trained, our objective is to maximize the expected quality of the generated images:
\begin{equation}
	\underset{{\bm{\phi}}}{\text{maximize}} \quad J({\bm{\phi}}) = \mathbb{E}_{\bm{x}\sim p_{\bm{\theta}}(\bm{x}\mid \bm{\eta}_{{\bm{\phi}}})} [r(\bm{x})],\label{eq:policy overall objective}
\end{equation}
where $r(\cdot)$ is a function quantifying image quality, as detailed in Section~\ref{sec:reward}, and $p_{\bm{\theta}}(\bm{x}\mid \bm{\eta}_{{\bm{\phi}}})$ denotes the distribution of the images generated by the generative model $\bm{\theta}$, under the generation policy specified by the policy network $\bm{\eta}_{\bm{\phi}}$.

It is challenging to solve problem \eqref{eq:policy overall objective} with straightforward gradient-based optimization for three reasons.
First, the function \( r(\cdot) \) that quantifies image quality is not necessarily differentiable.
Second, certain generative paradigms, such as MaskGIT and autoregressive models, inherently involve discrete operations (\eg, token sampling) that block gradient flow during backpropagation.
Third, even with a fully differentiable reward function and generation process, the iterative nature of modern generative models would render full-sequence gradient backpropagation prohibitively expensive.

\begin{table*}[t]
	\centering
	\caption{Unified MDP Formulation for Generative Models}
	\vspace{-5pt}
	\small
	\tablestyle{17.5pt}{.8}
	\begin{tabular}{lcccc}
	\toprule
	{Type} & {State $\bm{s}_t$} & {Action $\bm{a}_t$} & {Transition Distribution} $P(\bm{s}_{t+1}|\bm{s}_t,\bm{a}_t)$ & {Reward} $R(\bm{s}_t, \bm{a}_t)$ \\
	\midrule
	MaskGIT
	& $\big(t, \bm{x}_t\big)$
	& $\big(m_t, \tau_t, \zeta_t, w_t\big)$
	& $\Big(\delta(t+1), p_{\bm{\theta}}(\bm{x}_{t+1} | \bm{x}_t; \bm{a}_t)\Big)$
	& \multirow{8.3}{*}{$\mathbb{I}(t=T)\cdot r(\bm{x})$} \\[7pt]

	Autoregressive
	& $\big(t, \bm{x}_{1:t}\big)$
	& $\big(\tau_t, w_t, k_t, \rho_t\big)$
	& $\Big(\delta(t+1), p_{\bm{\theta}}(\bm{x}_{1:t+1}|\bm{x}_{1:t}; \bm{a}_t)\Big)$
	&  \\[7pt]

	Diffusion (ODE)
	& $\big(t, \bm{x}_{\kappa_t}\big)$
	& $\big(\kappa_{t+1}, w_t\big)$
	& $\Big(\delta(t+1), \delta\big(\text{ODE}(\bm{x}_{\kappa_t}, \kappa_{t+1}, \epsilon_{\bm{\theta}, w_t})\big)\Big)$
	& \\[7pt]

	Rectified Flow
	& $\big(t, \bm{x}_{\kappa_t}\big)$
	& $\big(\kappa_{t+1}, w_t\big)$
	& $\Big(\delta(t+1), \delta\big(\text{ODE}(\bm{x}_{\kappa_t}, \kappa_{t+1}, v_{\bm{\theta}, w_t})\big)\Big)$
	&  \\

	\bottomrule
	\end{tabular}
	\label{tab:mdp_unified}
	\begin{minipage}{.98\linewidth}
		\vspace{.4em}
		\scriptsize
		Here, $\delta(\cdot)$ is the Dirac delta function that is nonzero only at $\cdot$, representing a deterministic transition.
		$\text{ODE}(\cdot)$ encapsulates the numerical ODE solver step, while $\epsilon_{\bm{\theta}, w_t}$ and $v_{\bm{\theta}, w_t}$ are the noise and velocity prediction models respectively, guided by classifier-free guidance of scale $w_t$ (see Sec.~\ref{sec:prelim}).
		The reward definition $\mathbb{I}(t=T) \cdot r(\bm{x})$ indicates that rewards are only provided at the terminal step $T$, where $r(\cdot)$ (see Sec.~\ref{sec:reward}) evaluates the quality of the final generated image $\bm{x}$ extracted from the terminal state $\bm{s}_T$.
	\end{minipage}
	\vspace{-1em}
\end{table*}

\paragraph{Unified MDP Formulation for Generative Models.}
To address this issue, we propose to reformulate the determination of the proper generation policy as a Markov decision process (MDP).
Under this framework, the policy network $\bm{\eta}_{{\bm{\phi}}}$ can be naturally defined as an agent that observes the generation status and takes actions to configure the policy, which can then be optimized via reinforcement learning.
Our formulation, summarized in Table~\ref{tab:mdp_unified}, systematically characterizes diverse generative paradigms while preserving their distinctive characteristics.
Formally, we define a $T$-horizon MDP structure $(\mathcal{S}, \mathcal{A}, P, R)$ where:

\begin{itemize}[topsep=0pt]
	\item $\mathcal{S}$ denotes the state space.
	Each state $\bm{s}_t \in \mathcal{S}$ comprises dual components essential for adaptive policy decisions: 1) the current generation step $t$, signaling the generation progress in the finite $T$-horizon trajectory, and 2) the intermediate generation result at step $t$.
	For MaskGIT, the intermediate result is the partially masked token sequence $\bm{x}_t$.
	For autoregressive models, the intermediate result is the partial generation $\bm{x}_{1:t}$.
	For diffusion and rectified flow models, the intermediate result is the partially denoised sample $\bm{x}_{\kappa_t}$ at the ODE timestep $\kappa_t$.
	\item $\mathcal{A}$ denotes the action space. Each action $\bm{a}_t \in \mathcal{A}$ is defined as the specific generation policy required when transitioning from step $t$ to step $t+1$, which are already elaborated in Section~\ref{sec:prelim}.
	\item $P$ denotes the state transition probability function, which captures how states evolve under the influence of actions.
	While the generation step is deterministically updated from $t$ to $t+1$, the generation status evolves according to the underlying generative mechanism.
	For diffusion and rectified flow models, state transitions are deterministic and governed by numerical ODE solvers: $\bm{x}_{\kappa_{t+1}} = \text{ODE}(\bm{x}_{\kappa_t}, \kappa_{t+1}, \cdot)$, where $\cdot$ is either the noise prediction model $\epsilon_{\bm{\theta}, w_t}$ or velocity field model $v_{\bm{\theta}, w_t}$, guided by classifier-free guidance of scale $w_t$.
	For MaskGIT and autoregressive models, transitions follow stochastic conditional distributions $p_{\bm{\theta}}(\bm{x}_{t+1}|\bm{x}_t; \bm{a}_t)$ or $p_{\bm{\theta}}(\bm{x}_{1:t+1}|\bm{x}_{1:t}; \bm{a}_t)$ learned by the generative model.
	\item $R$ defines the reward function, which guides the policy network's behavior toward making better policy decisions.
	Consistent with our objective in Eq.~\eqref{eq:policy overall objective} that focuses on the final outcome, the reward is defined to be non-zero only at the terminal state, $t = T$:
	\begin{equation}
		R(\bm{s}_t, \bm{a}_t) \triangleq \mathbb{I}(t=T) \cdot r(\bm{x}),
	\end{equation}
	where $r(\cdot)$ (Section~\ref{sec:reward}) measures the quality of the generated image $\bm{x}$ extracted from the terminal state $\bm{s}_T$, with the extraction process involving obtaining the generated sample from the state tuple $\bm{s}_T$ and decoding to pixels when models operate in latent space~\cite{rombach2022high,peebles2023scalable}.
\end{itemize}

\begin{figure*}[t]
	\centering
	\includegraphics[width=\linewidth]{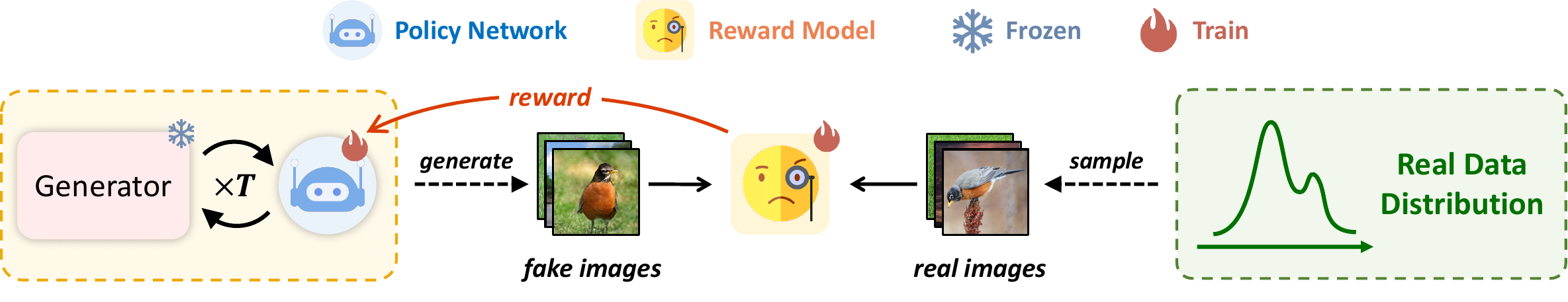}
	\vspace{-2.5em}
	\caption{
		\textbf{Training pipeline of \Ours with adversarial reward modeling.}
		The process involves an adversarial interplay: the \textcolor{agentcolor}{\textbf{policy network}} is optimized to maximize a reward signal, while the \textcolor{rmcolor}{\textbf{reward model}} is concurrently trained to distinguish real from generated images. The reward signal is the probability of a sample being deemed real by the reward model. Notably, the pre-trained generative model stays \textcolor{frozen}{\textbf{frozen}} throughout the pipeline.
	}
	\label{fig:pipeline}
	\vspace{-0.9em}
\end{figure*}
\begin{figure*}[t]
	\begin{center}
		\begin{minipage}{.55\linewidth}
			\includegraphics[width=\linewidth]{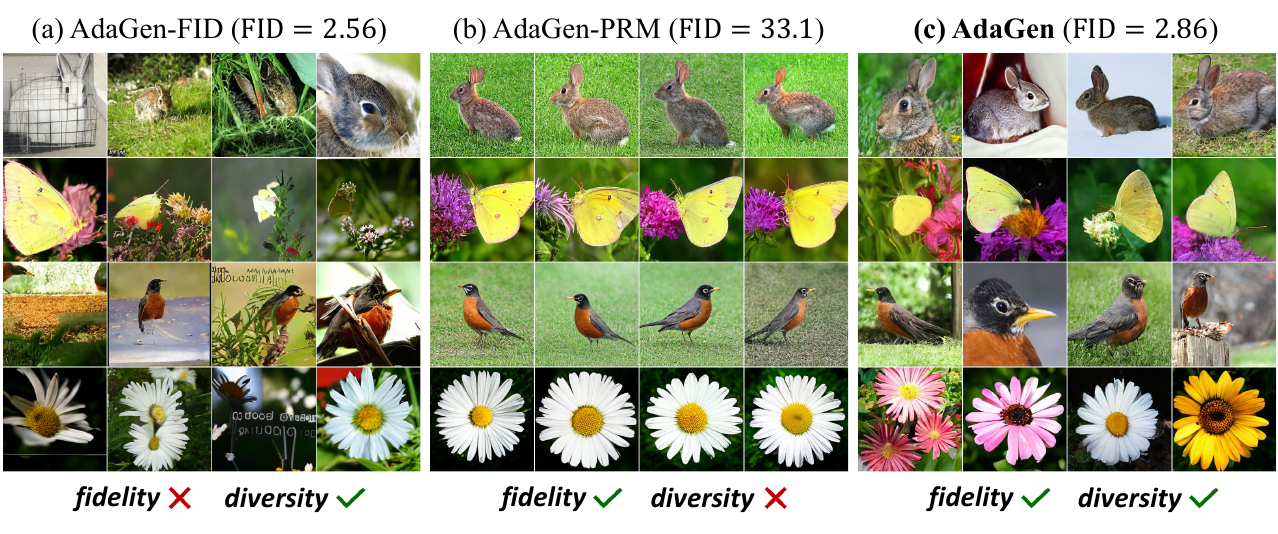}
			\vspace{-26pt}
			\caption{
				\textbf{Reward Design.\label{fig:comp_rewards}}
				(a) Using FID~\cite{salimans2016improved} as the reward.
				(b) Using a pre-trained reward model (PRM)~\cite{xu2024imagereward}.
				(c) Our main approach with adversarial reward modeling.
			}
		\end{minipage}
	\resizebox{.441\linewidth}{!}{
	\begin{minipage}[t]{\columnwidth}
		\vspace{-82pt}
		\begin{algorithm}[H]
			\caption{Training Procedure for \Ours \label{algo:ours}}
			\label{algo:ours_revised}
			\begin{algorithmic}[1]
				\Require Policy network $\bm{\pi}_{\bm{\phi}}$, reward model $r_{\bm{\psi}}$, generative model $\bm{\theta}$, real data distribution $p_{\text{data}}$
				\For{$i = 0, 1, 2, \dotsc$}
					\State \textcolor{blue}{\texttt{\# Policy network optimization}}
					\State Sample images $\bm{x}_{\text{fake}}\sim p_{\bm{\theta}}(\bm{x} \mid \bm{\pi}_{\bm{\phi}})$
					\State Update $\bm{\phi}$ with $\nabla_{\bm{\phi}} L^{\text{PPO}}(\bm{\phi})$ \Comment{See Eq.~\eqref{eq:ppo}}
					\State \textcolor{blue}{\texttt{\# Reward model optimization}}
					\State Sample images $\bm{x}_{\text{real}} \sim p_{\text{data}}$, $\bm{x}_{\text{fake}} \sim p_{\bm{\theta}}(\bm{x} \mid \bm{\pi}_{\bm{\phi}})$
					\State Update $\bm{\psi}$ with $\nabla_{\bm{\psi}} J(\bm{\phi}, \bm{\psi})$ \Comment{See Eq.~\eqref{eq:maximin_game_adv}}
				\EndFor
			\end{algorithmic}
		\end{algorithm}
	\end{minipage}
}
	\end{center}
	\vspace{-0.9em}
\end{figure*}

\paragraph{Learning the Adaptive Policy.}
Given our unified MDP formulation, we are able to formalize the policy network as a parametric agent $\bm{\eta}_{{\bm{\phi}}}$ that determines the proper generation policy $\bm{a}_t$ based on the current generation status $\bm{s}_t$:
\begin{equation}
    \bm{a}_t = \bm{\eta}_{\bm{\phi}}(\bm{s}_t).
\end{equation}
This formulation enables adaptive policy decisions tailored to each specific generation trajectory. To balance exploration and exploitation during policy learning, we implement the agent as a stochastic version of the deterministic policy network $\bm{\eta}_{{\bm{\phi}}}$:
\begin{equation}
    \bm{\pi}_{\bm{\phi}}(\bm{a}_t \mid \bm{s}_t) \triangleq \mathcal{N}\big(\bm{\eta}_{\bm{\phi}}(\bm{s}_t), \sigma \bm{I}\big),\label{eq:policy_to_agent}
\end{equation}
where $\mathcal{N}(\cdot)$ denotes a multivariate normal distribution, and $\sigma$ is a hyperparameter controlling the degree of exploration. During inference, we simply use the mean of this distribution, which reduces to the deterministic policy $\bm{\eta}_{\bm{\phi}}(\bm{s}_t)$.

We train $\bm{\pi}_{\bm{\phi}}$ to maximize the expected cumulative reward across the generation process:
\begin{equation}
    J({\bm{\phi}}) = \mathbb{E}_{\bm{\pi}_{\bm{\phi}}}\left[ \sum_{t=0}^{T} R(\bm{s}_t, \bm{a}_t) \right]= \mathbb{E}_{\bm{\pi}_{\bm{\phi}}}\big[R(\bm{s}_T, \bm{a}_T)\big],
\end{equation}
where the simplification to $R(\bm{s}_T, \bm{a}_T)$ follows from our terminal-only reward design. For effective policy gradient estimation, we adopt the Proximal Policy Optimization (PPO) algorithm~\cite{schulman2017proximal} with a clipped surrogate objective:
\begin{align}
    L^{PPO}({\bm{\phi}}) = \mathbb{E}_{t} \Big[ & \min\left( \rho_t({\bm{\phi}}) \hat{A}_t, \text{clip}\left(\rho_t({\bm{\phi}}), 1-\epsilon, 1+\epsilon\right) \hat{A}_t \right) \notag \\
    & - c \big( V_{{\bm{\phi}}}(\bm{s}_t) - R(\bm{s}_T, \bm{a}_T) \big)^2 \Big], \label{eq:ppo}
\end{align}
where $\rho_t({\bm{\phi}}) = \frac{\bm{\pi}_{\bm{\phi}}(\bm{a}_t \mid \bm{s}_t)}{\bm{\pi}_{{\bm{\phi}}_{\text{old}}}(\bm{a}_t \mid \bm{s}_t)}$ represents the probability ratio between new and old policies, $V_{\bm{\phi}}(\bm{s}_t)$ is a learned state-value function that shares the parameters with the policy network, $\hat{A}_t$ is the advantage estimate, and $\epsilon, c$ are hyperparameters. The advantage estimate is calculated as:
\begin{equation}
    \hat{A}_t = -V_{\bm{\phi}}(\bm{s}_t) + R(\bm{s}_T, \bm{a}_T).
\end{equation}
Further implementation details of our PPO optimization procedure are provided in Appendix~\ref{sec:imp_details}.

\subsection{Reward Design}
\label{sec:reward}

One of the key aspects in training our reinforcement learning agent is the design of the reward function.
In this section, we start with two straightforward reward designs and discuss their challenges.
Then, we propose an adversarial reward design that effectively addresses these challenges.

\hypertarget{opt:1}{}
\paragraph{1) Pre-defined Evaluation Metric.}
The most straightforward reward design is to employ commonly used evaluation metrics in image generation tasks such as Fr\'echet Inception Distance (FID).
However, we observe two challenges in practice.
\emph{First}, statistical metrics face challenges in providing sample-wise reward signals.
Evaluation metrics in image generation tasks are usually statistical, \ie, computed over a large number of samples.
For example, the common practice in the ImageNet 256$\times$256 benchmark is to evaluate the FID and IS score over 50K generated images~\cite{peebles2023scalable,bao2022all}, which makes it challenging to attribute the reward to specific actions.
In practice, we find this lack of informative feedback leads to failure in training the adaptive policy network, as detailed in Appendix~\ref{sec:app_exp_details}.
\emph{Second}, better metric scores do not necessarily translate into better visual quality.
As shown in Figure~\hyperref[fig:comp_rewards]{\ref*{fig:comp_rewards}a}, even when the evaluation metric FID is successfully optimized (\eg, using a non-adaptive variant of \Ours and achieving a low FID score of 2.56), the generated images may still suffer from poor visual quality.
This underscores the limitations of directly adopting evaluation metrics as optimization objectives and motivates us to find better alternatives for the reward function.

\hypertarget{opt:2}{}
\paragraph{2) Pre-trained Reward Model.}
Another option is to adopt a pre-trained, off-the-shelf reward model~\cite{xu2024imagereward} specialized at assessing the visual quality of images.
This strategy mitigates the two challenges in option~\hyperlink{opt:1}{1}, as it provides per-image reward signals that align more closely with perceptual quality.
However, we observe that the generated images in this case tend to converge on a similar style with relatively low diversity, as shown in Figure~\hyperref[fig:comp_rewards]{\ref*{fig:comp_rewards}b}.
Moreover, constructing such reward models typically requires large-scale human preference annotations and the training of additional deep networks~\cite{xu2024imagereward,xu2024visionreward}, which is generally costly.
Therefore, access to a pre-trained image reward model cannot be assumed in all scenarios.

\paragraph{\textbf{3) Adversarial Reward Modeling.}}
One shared issue with the aforementioned two designs is that, while the expected value of rewards can be effectively maximized, the resultant images exhibit unintended inferior quality or limited diversity.
In other words, the policy network tends to ``overfit'' these rewards.
Motivated by this observation, we hypothesize that a potential solution is to consider a reward that is updated concurrently with the policy network and adjusted dynamically to resist overfitting. 
To this end, we propose adversarial reward modeling, where we learn an adversarial reward model $r_{\bm{\psi}}$ together with the policy network.
Specifically, we formulate $r_{\bm{\psi}}$ as a discriminator akin to that in GANs~\cite{goodfellow2014generative} and establish a minimax game between the policy network and $r_{\bm{\psi}}$:
\begin{align}
	\underset{{\bm{\phi}}}{\text{max}}\ \underset{{\bm{\psi}}}{\text{min}}\ J({\bm{\phi}}, {\bm{\psi}}) =\
    &\mathbb{E}_{\bm{x}_{\text{fake}}\sim p_{\bm{\theta}}(\bm{x}\mid \bm{\pi}_{{\bm{\phi}}})} \bigl[ \log r_{\bm{\psi}}(\bm{x}_{\text{fake}}) \bigr] \notag \\
    \quad +\  &\mathbb{E}_{\bm{x}_{\text{real}}\sim p_{\text{data}}} \bigl[ \log \big(1 - r_{\bm{\psi}}(\bm{x}_{\text{real}})\big) \bigr],
	\label{eq:maximin_game_adv}
\end{align}
where $p_{\text{data}}$ denotes the distribution of real images.
As the policy tries to maximize the reward, the reward model is refined simultaneously to better distinguish between real and generated samples.
Consequently, we effectively curb the policy network from overfitting a static objective, and more balanced diversity and fidelity of the generated images are also observed (see Figure~\hyperref[fig:comp_rewards]{\ref*{fig:comp_rewards}c}).
Moreover, the adversarial reward model offers immediate, sample-wise reward signals, contrasting with the statistical metrics used in option \hyperlink{opt:1}{1}; and it obviates the need for expensive human preference data required by the pre-training of reward models in option \hyperlink{opt:2}{2}.
We summarize the training procedure for \Ours in Figure~\ref{fig:pipeline} and Algorithm~\ref{algo:ours}.

\subsection{Stabilizing the Exploration of Action Space}
\label{sec:action_stablize}
\begin{figure}[t!]\centering
\includegraphics[width=\linewidth]{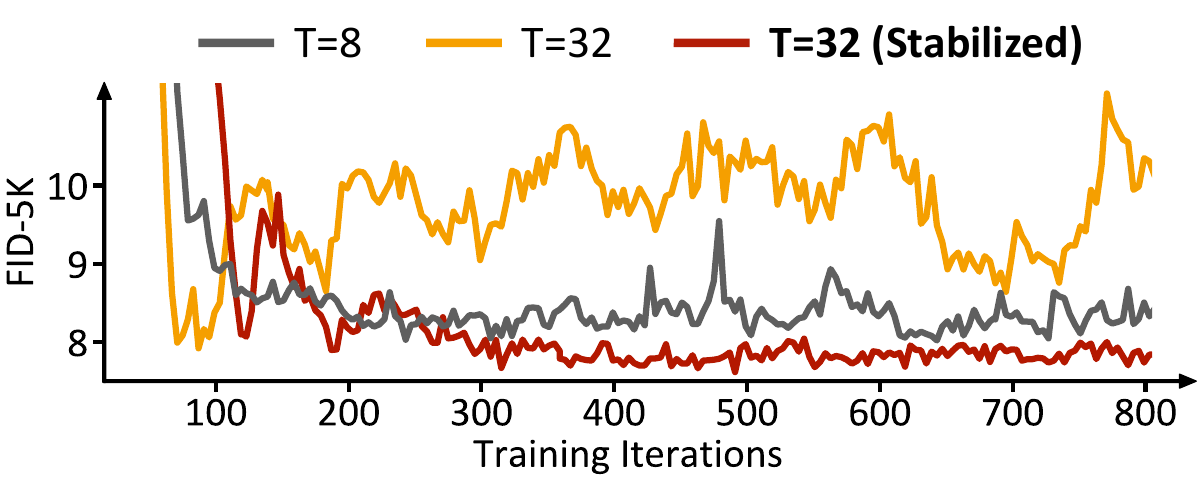}
\vspace{-2em}
\caption{
	\textbf{Optimization process of \Ours.}
	When the generation steps increase from	$T=8$ to $T=32$, the optimization becomes unstable and yields worse performance.
	Our proposed action smoothing technique stabilizes convergence and achieves better performance ($T=32$ (Stabilized)).
}
\vspace{-0.9em}
\label{fig:momentum_abl}
\end{figure}

Our framework requires the policy network to operate in a high-dimensional action space, producing multiple policy parameters across generation steps to maximize rewards. While Algorithm~\ref{algo:ours} effectively mitigates reward overfitting through the joint optimization of the reward and policy networks, we observe that training becomes increasingly unstable as the action space expands. As shown in Figure~\ref{fig:momentum_abl}, increasing the number of generation steps from $T = 8$ to $T = 32$ in MaskGIT, which effectively expands the action space by 4 times, leads to greater instability without performance gains.

\begin{figure}[h!]
	\centering
	\includegraphics[width=\columnwidth]{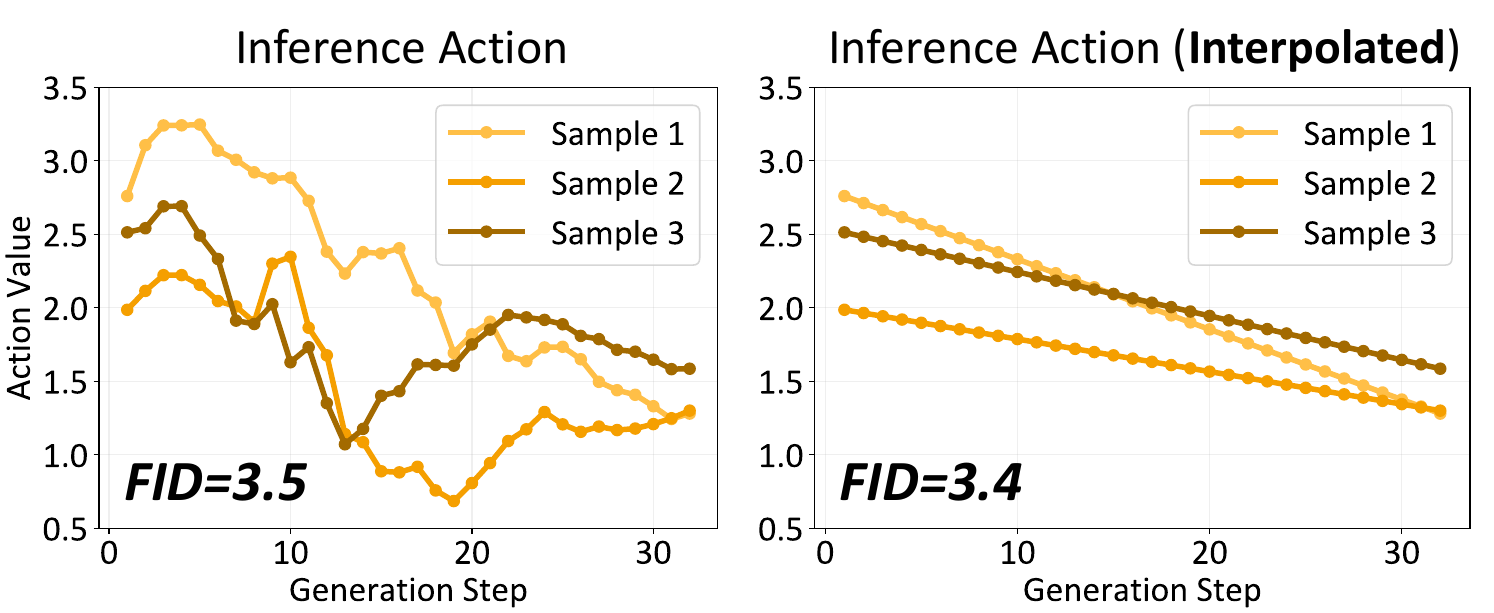}
	\vspace{-2em}
	\caption{
		\textbf{Action sequence analysis}.
		\emph{Left}: Learned action sequences at $T = 32$ for three random samples exhibit erratic fluctuations.
		\emph{Right}: A simple linear interpolation between the start and end points of the original learned action sequence yields slightly better performance.
		We visualize sampling temperature $\tau_t$ as an example of the action sequence.
		\label{fig:inspect_t32_action}
	}
	\vspace{-1.1em}
\end{figure}

\begin{figure}[h!]
 \centering
 \includegraphics[width=\columnwidth]{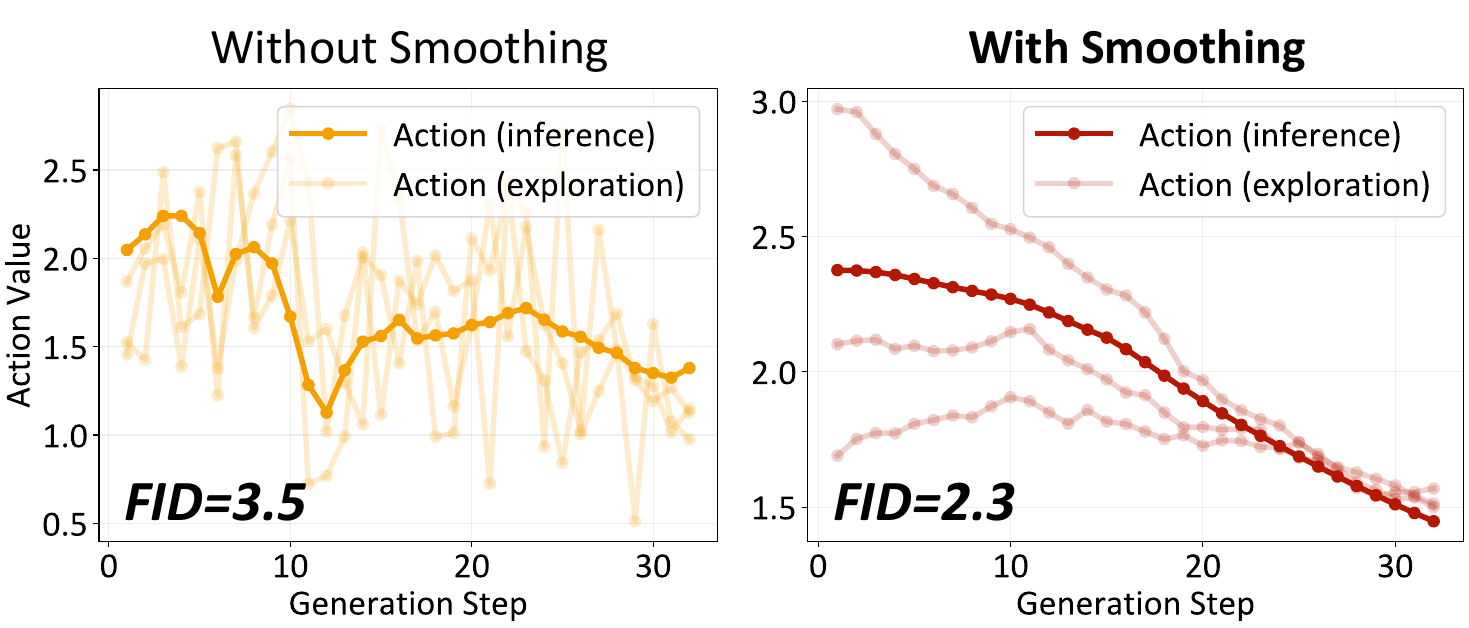}
 \vspace{-2em}
 \caption{
	\textbf{Comparison of action sequences without (left) and with (right) action smoothing.}
	We visualize sequences from both training (exploration) and inference.
	Action smoothing leads to more feasible exploration during training, and the inference action sequence is also much more reasonable.
 }
\vspace{-0.9em}
 \label{fig:stability}
\end{figure}

\paragraph{Inspecting the Instability Issue.}
To investigate this issue, we examined the policy's action sequences at $T = 32$ and observed erratic fluctuations (Figure~\ref{fig:inspect_t32_action}, left).
This raises the question of whether such fluctuations are truly necessary for effective generation.
To test this, we conducted a controlled experiment in which the learned action sequences were replaced with a simple linear interpolation between their initial and final values.
Interestingly, this naive approach even slightly improved the performance, as shown in Figure~\ref{fig:inspect_t32_action} (right).
This suggests that the high-frequency variations captured by the original policy network may be not only unnecessary but also potentially harmful to performance.
This observation is consistent with the characteristics of iterative, progressive generation paradigms:
Since these models typically begin with high uncertainty and gradually reveal more information, it is less likely that the optimal policy would require highly non-monotonic behavior with irregular fluctuations.

Given that these fluctuations appear unnecessary for effective generation, we further investigate why they occur in the first place.
The phenomenon may arise from the unstructured exploration process described in Eq.~\eqref{eq:policy_to_agent}, where Gaussian noise with covariance $\sigma \bm{I}$ is independently added to the policy network's output at each time step.
This approach generates highly unstructured exploration trajectories (Figure~\ref{fig:stability}, left).
However, as noted above, such action sequences are rarely suitable for progressive generation and thus may lead to inefficient exploration.
Meanwhile, previous studies have also shown that this type of unstructured exploration tends to produce noisy, high-variance inference policy trajectories~\cite{kober2008policy, raffin2022smooth}, which aligns with the fluctuations observed in Figure~\ref{fig:inspect_t32_action}.

\paragraph{Stabilizing Exploration via Action Smoothing.}
To enhance exploration stability, we introduce a simple yet powerful technique: \emph{action smoothing}. The core idea is to smooth the policy network's raw output sequence before execution. This process suppresses high-frequency oscillations, thereby mitigating erratic, high-variance trajectories and fostering more stable and efficient policy optimization.

Formally, we consider a transformation function \( \bm{f} \) that maps the raw output sequence \( \tilde{\bm{a}}_{1:T} \) sampled from the policy network \( \bm{\pi}_{\bm{\phi}} \) to the final, smoothed action sequence \( \bm{a}_{1:T} \):
\begin{equation}
	\bm{a}_{1:T} = \bm{f}(\tilde{\bm{a}}_{1:T}), \label{eq:final_action}
\end{equation}
with two requirements imposed on the transformation \( \bm{f} \):
\begin{itemize}
	\item \textbf{Low-pass filtering}: The function should suppress high-frequency fluctuations to produce smoother action sequences with fewer oscillations.
	\item \textbf{Causality}: The transformation must satisfy the causality constraint, ensuring that the action output $\bm{a}_t$ at time $t$ must not depend on any future inputs $\tilde{\bm{a}}_{i}$ for $i > t$. This is essential to preserve the Markovian property of our MDP formulation (Section~\ref{sec:ours}).
\end{itemize}

In theory, any causal low-pass filter from signal processing can serve as a valid choice for \( \bm{f} \). Among these, we find that the simplest single-pole low-pass IIR (Infinite Impulse Response) filter, commonly known as an Exponential Moving Average (EMA) filter~\cite{roberts1987digital}, is already highly effective:
\begin{equation}
	\bm{a}_t = \beta \bm{a}_{t-1} + (1 - \beta) \tilde{\bm{a}}_t, \label{eq:define_momentum}
\end{equation}
where \( \beta \in [0, 1] \) is a hyperparameter that controls the degree of smoothing. This straightforward approach effectively smooths the $T=32$ action sequences (Figure~\ref{fig:stability}), leading to more stable training and a clear performance improvement over the \( T = 8 \) baseline (Figure~\ref{fig:momentum_abl}).
We present the ablation study on the action smoothing weight $\beta$ in Table~\ref{tab:abl_momentum}.

\subsection{Improvements of \Ours}
\label{sec:improvements}

\subsubsection{Inference-time Refinement}
\label{sec:inference_refinement}
Our framework uses an adversarial reward model \( r_{\bm{\psi}} \) and a value network \( V_{\bm{\phi}} \) during training to provide learning signals and improve sample efficiency.
However, they are originally discarded after training and only the policy network is retained to generate outputs at inference time.
In this section, we demonstrate that these two auxiliary networks can be repurposed at inference time to further enhance the generation performance.

\begin{figure}[h!]
    \centering
    \includegraphics[width=\columnwidth]{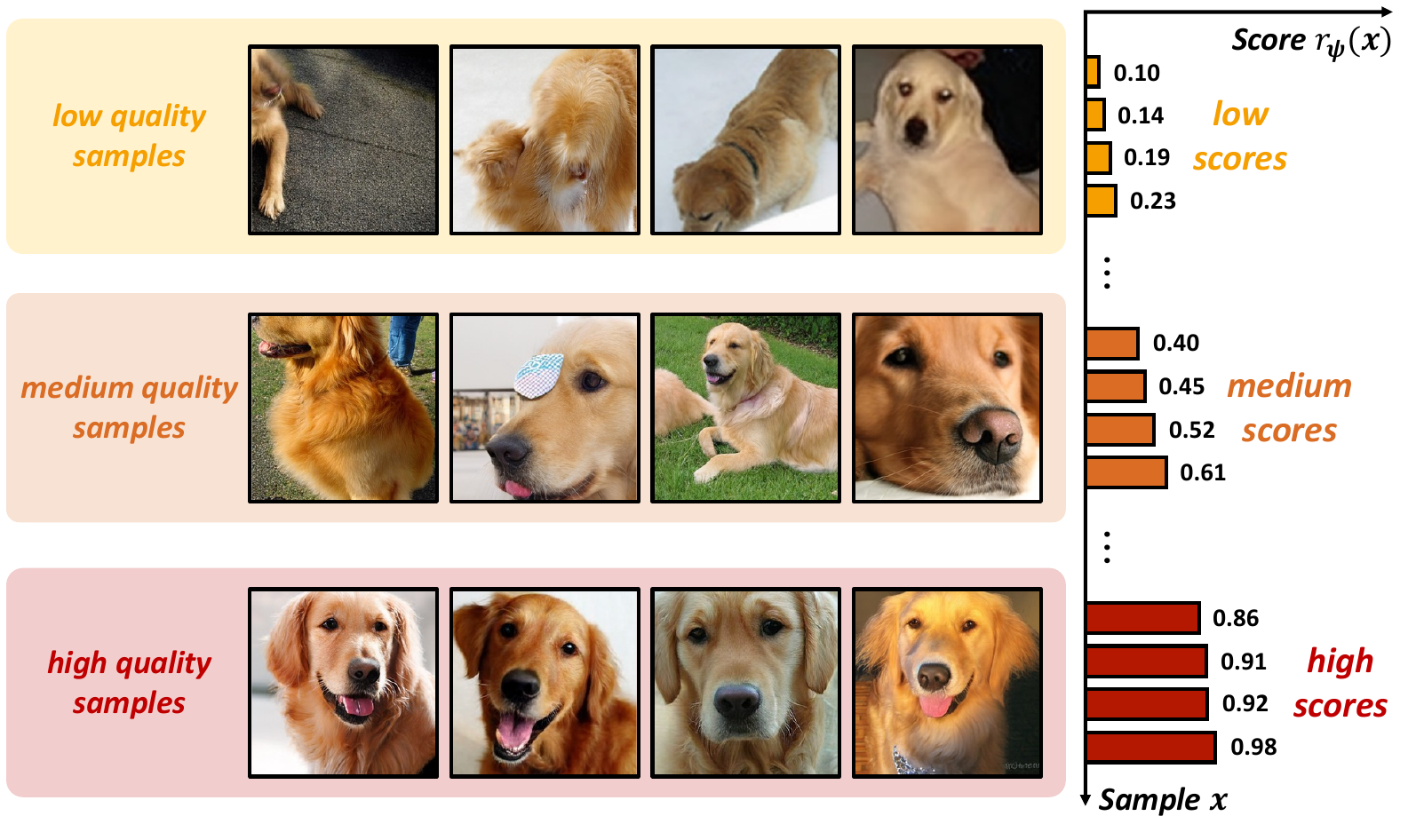}
	\vspace{-2.4em}
    \caption{
		\textbf{Generated samples and their scores from our adversarial reward model}. We showcase 12 samples and observe that the assigned scores, $r_{\bm{\psi}}(\bm{x})$, align well with their visual quality. This demonstrates that our adversarial reward model serves as an effective evaluator for perceptual quality.
		\label{fig:img_with_score}
    }
	\vspace{-1.1em}
\end{figure}

First, we leverage the adversarial reward model, which has been exposed to a large volume of both real and generator-produced images during training.
We observe that its score can serve as an effective proxy for the quality of a generated image, as illustrated in Figure~\ref{fig:img_with_score}.
Therefore, we use this score during inference to guide repeated sampling~\cite{brown2024large,ma2025inference}, selecting the generation with the highest reward to enhance the final output.

Beyond repeated sampling based on the final reward, we further propose a lookahead sampling mechanism for mid-process refinement when the transition function $P(\bm{s}_{t+1}|\bm{s}_t,\bm{a}_t)$ is non-deterministic (\eg, MaskGIT, see Table~\ref{tab:mdp_unified} for more details).
Recall that the value network \( V_{\bm{\phi}} \) is trained to predict the expected final reward from an intermediate generation state $\bm{s}_t$ (see Eq.~\eqref{eq:ppo}). 
This property makes it well-suited for guiding local decision-making to improve sampling quality within a single generation process.
Specifically, at each transition from state $\bm{s}_t$ to $\bm{s}_{t+1}$, when the transition is stochastic, we can sample $K$ candidates $(\bm{s}_{t+1}^{(1)}, \dots, \bm{s}_{t+1}^{(K)})$, and then compute the predicted expected reward \( V_{\bm{\phi}}(\bm{s}_{t+1}^{(k)}) \) for each $k \in \{1, \dots, K\}$. We then select the next state with the highest expected reward:
\begin{equation}
    k^* = \arg\max_{k} V_{\bm{\phi}}(\bm{s}_{t+1}^{(k)}), \quad \bm{s}_{t+1} = \bm{s}_{t+1}^{(k^*)}. \label{eq:lookahead_selection}
\end{equation}
This lookahead mechanism steers the generation process toward more promising trajectories, enabling within-process refinement.

We summarize our full inference-time refinement strategy in Algorithm~\ref{algo:inference_refinement}. The algorithm includes both the repeated sampling and the optional lookahead sampling.

\begin{algorithm}[htbp]
    \caption{Inference-Time Refinement for \Ours \label{algo:inference_refinement}}
    \begin{algorithmic}[1]
        \Require Policy network $\bm{\eta}_{\bm{\phi}}$, adversarial reward model $r_{\bm{\psi}}$, value network $V_{\bm{\phi}}$. Number of repeated sampling trials $M$. Number of lookahead branches $K$.
        \State Initialize $\bm{x}_{\text{best}} \leftarrow \text{null}$, $r_{\text{best}} \leftarrow -\infty$
        \For{$i = 0$ to $M$}  \Comment{Repeated Sampling}
			\State Initialize generation state $\bm{s}_0$
            \For{$t = 0$ to $T-1$}
				\State Obtain action: $\bm{a}_t \leftarrow \bm{\eta}_{\bm{\phi}}(\bm{s}_t)$
				\If{Lookahead}  \Comment{Lookahead Sampling}
					\State Sample: $\{\bm{s}_{t+1}^{(k)}\}_{k=1}^K \sim P(\bm{s}_{t+1}|\bm{s}_t,\bm{a}_t)$
					\State Find optimal index: $k^* \leftarrow \arg\max_{k} V_{\bm{\phi}}(\bm{s}_{t+1}^{(k)})$
					\State Select optimal state: $\bm{s}_{t+1} \leftarrow \bm{s}_{t+1}^{(k^*)}$
				\Else
					\State Sample: $\bm{s}_{t+1} \sim P(\bm{s}_{t+1}|\bm{s}_t,\bm{a}_t)$
				\EndIf
            \EndFor
			\State Extract final image $\bm{x}$ from the terminal state $\bm{s}_T$
            \State Compute reward: $r \leftarrow r_{\bm{\psi}}(\bm{x})$
            \If{$r > r_{\text{best}}$}
                \State $r_{\text{best}} \leftarrow r$
                \State $\bm{x}_{\text{best}} \leftarrow \bm{x}$
            \EndIf
        \EndFor
        \State \Return $\bm{x}_{\text{best}}$
    \end{algorithmic}
\end{algorithm}

\subsubsection{Flexibly Trading Off Fidelity and Diversity}
\label{sec:tradeoff}
While our adversarial reward model is designed to achieve a strong, balanced trade-off of fidelity and diversity, certain applications may demand prioritizing one metric over the other. For instance, a user might be willing to sacrifice some output variety in exchange for the highest possible image fidelity. To accommodate such requirements, we introduce a mechanism that offers explicit, granular control over the fidelity-diversity trade-off, governed by a user-adjustable parameter, $\lambda$.

\begin{figure}[t!]
	\centering
	\includegraphics[width=.43\textwidth]{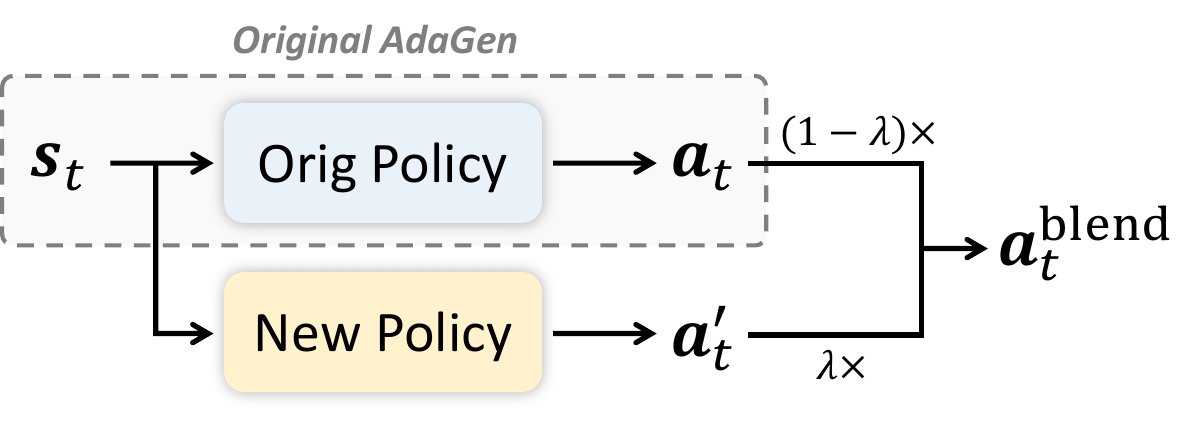}
	\vspace{-1.1em}
	\caption{
	   \textbf{Illustration of our blended policy.}
	   We introduce a new, fidelity-oriented policy network upon our original framework and linearly interpolate its output $\bm{a}_t'$ with the original policy network's output $\bm{a}_t$ with a blending parameter $\lambda$ to obtain the final action $\bm{a}_t^{\rm blend}$.
	   \label{fig:tradeoff}
	}
	\vspace{-1em}
\end{figure}

As illustrated in Figure~\ref{fig:tradeoff}, our strategy utilizes two policy networks. The first is our original policy network trained with adversarial reward modeling. We then introduce a secondary policy network that solely focuses on image fidelity.
The final action $\bm{a}_t^{\rm blend}$ is derived from a linear interpolation of the action $\bm{a}_t$ from our primary policy and the action $\bm{a}_t'$ from the fidelity-oriented policy:
\begin{equation}
 \bm{a}_t^{\rm blend} = (1 - \lambda) \, \bm{a}_t + \lambda \, \bm{a}_t'. \label{eq:tradeoff_action}
\end{equation}
Here, $\lambda \in [0, 1]$ is a user-adjustable hyperparameter that allows a smooth transition from a balanced synthesis at $\lambda = 0$ to outputs of maximal fidelity as $\lambda$ approaches 1.

The training of the fidelity-oriented policy network is governed by a dynamic objective. In each training step, we sample a blending factor $\lambda \sim \mathcal{U}[0, 1]$ and generate an image using the interpolated policy. The fidelity-oriented policy network is then optimized with a blended reward signal:
\begin{equation}
 r_{\rm blend} = (1 - \lambda) \, r + \lambda \, r',
\end{equation}
while the original policy network is kept frozen to maintain the integrity of the balanced policy.
In this formulation, $r$ is our original adversarial reward, whereas $r'$ is a dedicated fidelity reward, for which we employ ImageReward~\cite{xu2024imagereward}.
The key to this design is the dual function of $\lambda$; it controls both the fidelity-oriented policy's influence on the output and the weight of the fidelity-centric reward in its optimization objective.
This coupling establishes a explicit mapping between the control parameter $\lambda$ and the fidelity-diversity spectrum, thereby offering a user-controllable modulation of generation behavior.
The effectiveness of this blending strategy is demonstrated in Figure~\ref{fig:tradeoff_fid_div}.

\begin{table*}[!t]
	\centering
	\caption{
		{Effectiveness of \Ours on Different Generative Paradigms}\label{tab:fid_add_adagen}
	}
	\vspace{-.6em}
	\tablestyle{17pt}{1}
	\begin{tabular}{lccllll}
		\toprule
		Model & Type & Policy &$T=4$ & $T=8$ & $T=10$ & $T=16$ \\
		\midrule
		\multirow{2}{*}{MaskGIT-S} & \multirow{2}{*}{Mask.} & Baseline & 7.65 & 5.01 & - & 4.88 \\
		&  & {\textbf{\Ours}} & {\dt{4.54}{-3.11}} & {\dt{3.71}{-1.30}} & {-} & {\dt{3.36}{-1.52}} \\\midrule
		\multirow{2}{*}{MaskGIT-L} & \multirow{2}{*}{Mask.} & Baseline & 6.91 & 4.65 & - & 3.79 \\
		&  & {\textbf{\Ours}} & {\dt{3.63}{-3.28}} & {\dt{2.86}{-1.79}} & {-} & {\dt{2.41}{-1.38}} \\\midrule
		\multirow{2}{*}{DiT-XL} & \multirow{2}{*}{Diff.} & Baseline & 9.71 & 5.18 & - & 3.31 \\
		&  & {\textbf{\Ours}} & {\dt{5.31}{-4.40}} & {\dt{2.82}{-2.36}} & {-} & {\dt{2.19}{-1.12}} \\\midrule
		\multirow{2}{*}{SiT-XL} & \multirow{2}{*}{Flow} & Baseline & 9.33 & 4.90 & - & 2.99 \\
		&  & {\textbf{\Ours}} & {\dt{4.25}{-5.08}} & {\dt{2.72}{-2.18}} & {-} & {\dt{2.12}{-0.87}} \\\midrule
		\multirow{2}{*}{VAR-$d$16} & \multirow{2}{*}{AR} & Baseline & - & - & 3.30 & - \\
		&  & {\textbf{\Ours}} & {-} & {-} & {\dt{2.62}{-0.68}} & {-} \\\midrule
		\multirow{2}{*}{VAR-$d$30} & \multirow{2}{*}{AR} & Baseline & - & - & 1.92 & - \\
		&  & {\textbf{\Ours}} & {-} & {-} & {\dt{1.59}{-0.33}} & {-} \\
		\bottomrule
	\end{tabular}
	\begin{minipage}{.86\linewidth}
		\vspace{.4em}
		\scriptsize
		We evaluate \Ours on ImageNet 256$\times$256 using six models spanning four representative paradigms and report FID-50K results.
		As VAR uses a fixed number of 10 generation steps~\cite{Tian2024VisualAM}, we report results under the same configuration.
		Mask.: MaskGIT, Diff.: diffusion, Flow: rectified-flow, AR: autoregressive.
	\end{minipage}
	\vspace{-1.2em}
\end{table*}

\begin{figure*}[t!]
	\centering
	\includegraphics[width=.88\linewidth]{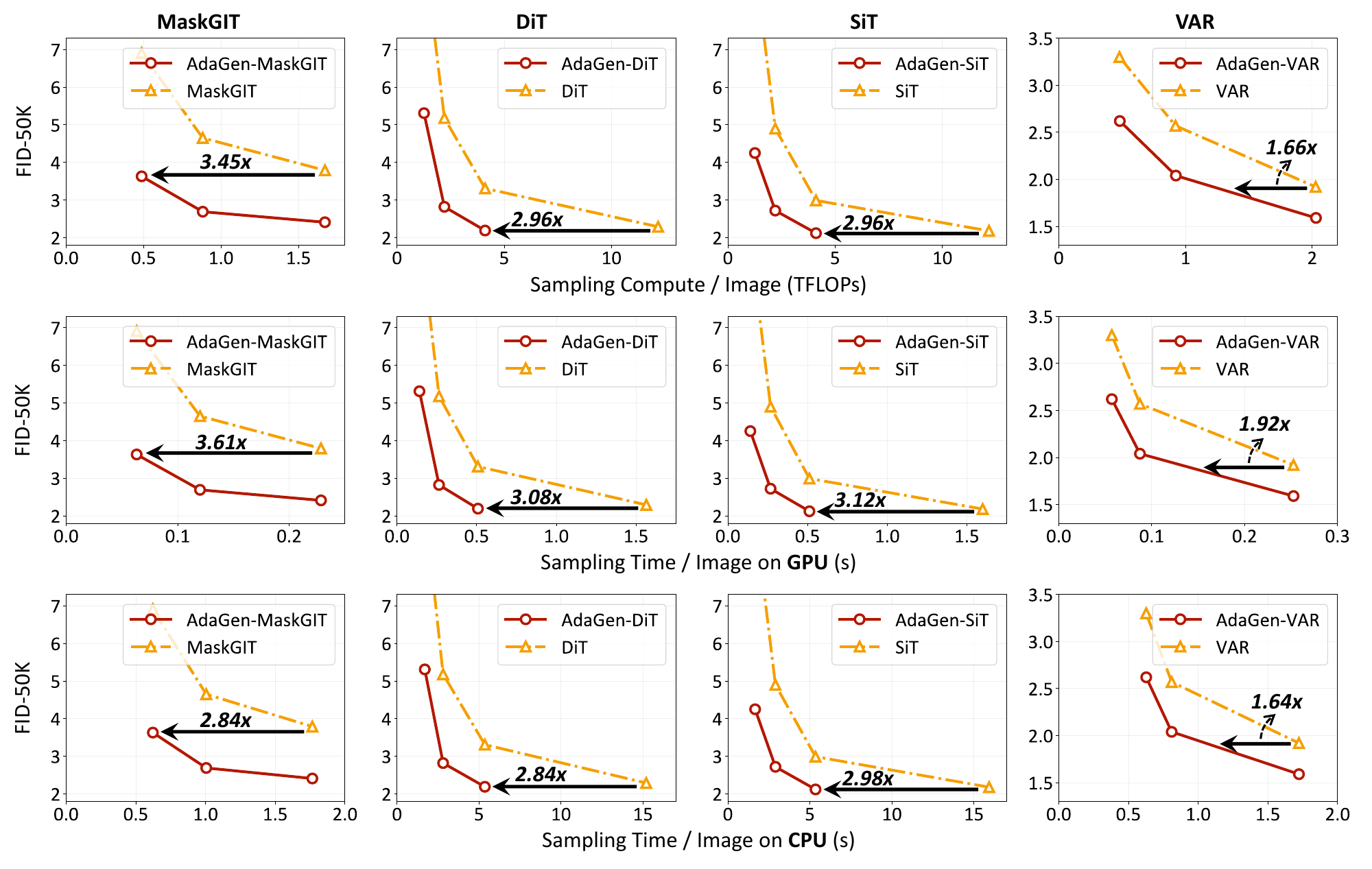}
	\vspace{-1.5em}
    \caption{\textbf{Efficiency of \Ours}.
			We vary the inference budget by adjusting the number of inference steps ($T=4, 8, 16$) for MaskGIT-L, SiT-XL, and DiT-XL. For VAR, whose inference steps are fixed, we vary the model size (VAR-$d$16, VAR-$d$20, VAR-$d$30) instead.
			GPU time is measured on an A100 GPU with batch size 16. CPU time is measured on Xeon 8375 CPU with batch size 1.
	}
    \label{fig:practical_speed}
\end{figure*}

\section{Experiments}
\label{sec:exp}

We conduct extensive experiments to demonstrate the effectiveness of \Ours. Section~\ref{sec:cin_exp} presents class-conditional generation on ImageNet with different architectures and generation steps. Section~\ref{sec:t2i_exp} covers text-to-image generation results on MS-COCO, CC3M, and LAION-5B. Sections~\ref{sec:inference_refinement_exp} and~\ref{sec:tradeoff_exp} examine \Ours's inference-time refinement capabilities and flexibility in balancing fidelity and diversity. Section~\ref{sec:analysis} provides comprehensive analyses of \Ours.

\subsection{Setups}
\paragraph{Datasets.} For class-conditional image generation, we use ImageNet~\cite{russakovsky2015imagenet} at 256$\times$256 and 512$\times$512 resolutions, reporting FID-50K following \cite{peebles2023scalable}. For text-to-image generation, we use MS-COCO~\cite{lin2014microsoft}, CC3M~\cite{sharma2018conceptual}, and LAION-5B~\cite{schuhmann2022laion}, reporting FID-30K following \cite{bao2022all,chang2023muse}. For large-scale text-to-image generation on LAION-5B, we follow common practice \cite{saharia2022photorealistic,chang2023muse} to report zero-shot FID-30K results on MS-COCO.

\paragraph{Generative Models.} We evaluate \Ours{} across four representative generative paradigms. For MaskGIT, we utilize the publicly available implementation from \cite{Ni2024AdaNAT} and evaluate MaskGIT-S with 13 layers, 512 dimensions, and approximately 50M parameters, and MaskGIT-L with 25 layers, 768 dimensions, and approximately 200M parameters. For diffusion models, we adopt DiT-XL/2~\cite{peebles2023scalable} for class-conditional generation on ImageNet~\cite{russakovsky2015imagenet} and Stable Diffusion~\cite{rombach2022high} for text-to-image generation on LAION-5B~\cite{schuhmann2022laion}. For rectified-flow models, we adopt SiT-XL~\cite{Ma2024SiTEF}. For autoregressive models, we employ VAR~\cite{Tian2024VisualAM} with different model scales.
Following recent trends in advanced samplers for diffusion and rectified-flow models~\cite{lu2022dpm,lu2022dpmp,Xie2024SANAEH}, we adopt DPM-Solver~\cite{lu2022dpm} for DiT-XL and SiT-XL, building \Ours{} upon it. For fair comparisons, we employ the same few-step samplers for the corresponding baselines. We also experiment with different generation steps. For models with more than 10 steps, we apply action smoothing with $\beta=0.8$ to stabilize policy network exploration (see Section~\ref{sec:action_stablize} for more details).

\paragraph{Implementation Details.}
In practice, our adaptive policy network leverages off-the-shelf features from the backbone generative models rather than raw intermediate generation results (e.g., $\bm{x}_t$) as input, which we find more effective. The architecture consists of a convolutional layer, a multi-layer perceptron (MLP), and additional adaptive LayerNorm (AdaLN) layers to incorporate generation step information. To ensure output policy parameters remain within appropriate ranges, we introduce activation functions, \eg, softplus to constrain the classifier-free guidance scale to be within $(0, +\infty)$, at the end of the policy network. The adversarial reward model architecture adopts a transformer-based discriminator following~\cite{sauer2023stylegan}. We present in-depth ablation studies for these architectural decisions in Table~\ref{tab:ablations}. 
Further implementation details can be found in Appendix~\ref{sec:imp_details}.

\subsection{Class-conditional Generation on ImageNet.}
\label{sec:cin_exp}

\paragraph{Effectiveness of \Ours on Different Generative Paradigms.}
Table~\ref{tab:fid_add_adagen} evaluates \Ours on ImageNet 256$\times$256 across six models spanning four iterative generation paradigms. For each paradigm except VAR with a fixed 10-step configuration, we test at 4, 8, and 16 inference steps.
Our adaptive policy consistently outperforms the corresponding baselines across all paradigms and inference budgets, with performance gains becoming more pronounced as inference steps decrease. With the DiT-XL backbone, our policy reduces FID by 1.12 at 16 steps, 2.36 at 8 steps, and 4.40 at 4 steps, demonstrating strong advantages in computationally constrained scenarios.

Figure~\ref{fig:practical_speed} illustrates generation quality-efficiency trade-offs across different models. For most models, we vary inference steps to represent different computational budgets. For VAR, which utilizes fixed generation steps, we vary the model capacity to represent different computational budgets. We report both theoretical inference cost measured by TFLOPs and practical GPU/CPU latency. Results demonstrate that \Ours consistently improves the quality-efficiency frontier, achieving 1.6$\times$ to 3.6$\times$ speedup while maintaining comparable generation quality. This enables reduced-budget inference to achieve competitive performance.

\paragraph{System-level Comparison.}
Tables~\ref{tab:fid_imnet} and~\ref{tab:fid_imnet512} compare \Ours-augmented models with strong baselines on ImageNet at 256$\times$256 and 512$\times$512 resolutions. Across all settings, \Ours consistently improves both performance and efficiency. For instance, \Ours-DiT-XL achieves an FID of 2.19 in 16 steps (4.1 TFLOPs), outperforming the baseline DiT-XL's 2.29 FID with 50 steps (12.2 TFLOPs), resulting in a $\sim3\times$ reduction in inference cost. On VAR-$d$30, \Ours improves the FID from 1.92 to 1.59 with negligible overhead. At 512$\times$512 resolution, \Ours-MaskGIT achieves 2.82 FID with 2.0 TFLOPs and 2.46 FID with 3.6 TFLOPs, outperforming baselines while remaining more efficient.

\begin{table*}[!t]
    \centering
    \caption{
       {Class-conditional Image Generation on ImageNet 256$\times$256}
       \label{tab:fid_imnet}
    }
    \vspace{-.5em}
    \tablestyle{10pt}{1}
    \resizebox{.9\linewidth}{!}{
    \begin{tabular}{y{140}x{30}x{46}x{27}x{45}x{41}x{20}}
       \toprule
       Method & Type &  Params & Steps & TFLOPs$\downarrow$ & FID-50K$\downarrow$ & IS$\uparrow$  \\
       \midrule
       \VQVAE             & AR    & 13.5B & 5120    & -    & 31.1  & $\sim$ 45    \\
       \VQGAN             & AR    & 1.4B  & 256     & -    & 15.78 & 78.3    \\
       ADM-G~\cite{dhariwal2021diffusion}\pub{NeurIPS'21}               & Diff. & 554M  & 250     & 334.0  & 4.59  & 186.7  \\
       ADM-G, ADM-U~\cite{dhariwal2021diffusion}\pub{NeurIPS'21} & Diff. & 608M  & 250 & 239.5    & 3.94  & 215.8    \\
       \LDM               & Diff. & 400M  & 250     & 52.3 & 3.60  & 247.7      \\
       \VQDiffusion       & Diff. & 554M  & 100     & 12.4 & 11.89  & -          \\
        \VITVQGAN & AR & 599M & 1024 & - & 3.04 & 227.4 \\
        \DraftRevise       & Mask.   & 1.4B  & 72      & -    & 3.41  & 224.6     \\
        \UVIT & Diff. & 501M & 50 & 13.6 & 2.29 & 263.8 \\
        \DIT & Diff. & 675M & 250 & 59.6 & 2.27 & 278.2  \\
        \MDT &  Mask. & 676M & 250 & 59.6 & 1.79 & 283.0 \\
        \SIMPLEDIFFUSION & Diff. & 2B & 512 & - & 2.44 & 256.3 \\
        \RIN & Diff. & 410M & 1000 & 334.0 & 3.42 & 182.0 \\
        \VDMPP & Diff. & 2B & 512 & - & 2.12 & 267.7 \\
        \SIT & Flow & 675M & 250 & 59.6 & 2.06 & 270.3  \\
       {MAGVIT-v2~\cite{yu2023language}\pub{ICLR'24}\xspace} & Mask. & 307M & 64 & - & 1.78 & 319.4 \\
       TiTok~\cite{yu2024image}\pub{NeurIPS'24}\xspace & Mask. & 287M & 64 & 4.3 & 1.97 & - \\
        \LLAMAGEN & AR & 3.1B & 256 & - & 2.18 & 263.3 \\
       \midrule
       {MaskGIT~\cite{chang2022maskgit}\pub{CVPR'22}} & {Mask.} & {227M} & {8} & {0.6} & {6.18} & {182.1} \\
       {MaskGIT$^\dagger$~\cite{chang2022maskgit}\pub{CVPR'22}} & {Mask.} & {227M} & {8} & {8.4} & {4.02} & {-} \\
       \multirow{2}{*}{\DIT} & Diff. & 675M & 16 & 4.1 & 3.31 & -  \\
       & Diff. & 675M & 50 & 12.2 & 2.29 &  275.3  \\
       \multirow{2}{*}{\SIT} & Diff. & 675M & 16 & 4.1 & 2.99 & -  \\
       & Diff. & 675M & 50 & 12.2 & 2.18 &  287.0 \\

       \TokenCritic & Mask. & 422M & 36 & 1.9 & 4.69 & 174.5 \\
       {Token-Critic$^\dagger$~\cite{lezama2022improved}\pub{ECCV'22}} & {Mask.} & {422M} & {36} & {8.7} & {3.75} & {287.0}  \\
       \MAGE & Mask. & 230M & 20 & 1.0 & 6.93 & -   \\
       \MaskgitFSQ & Mask. & 225M & 12 & 1.4 & 4.53 & - \\
        \USF & Diff. & 554M & 8 & 10.7 & 9.72 & - \\
        \RCGL & Mask. & 502M & 20 & 4.4 & 2.15 & 253.4 \\
       VAR-$d$16~\cite{Tian2024VisualAM}\pub{NeurIPS'24}\xspace & AR & 310M & 10 & 0.5 & 3.30 & 274.4 \\
       VAR-$d$30~\cite{Tian2024VisualAM}\pub{NeurIPS'24}\xspace & AR & 2.0B & 10 & 2.0 & 1.92 & 323.1 \\
       \midrule
       \textbf{\Ours-MaskGIT-L} & Mask. & 206M & 16 & 1.7 & 2.41 & 269.7 \\  %
       \textbf{\Ours-DiT-XL} & Diff. & 688M & 16 & 4.1 & 2.19 & 279.8 \\
       \textbf{\Ours-SiT-XL} & Flow & 688M & 16 & 4.1 & 2.12 & 291.4 \\
       \textbf{\Ours-VAR-$d$16} & AR & 323M & 10 & \textbf{0.5} & 2.62 & 278.4 \\
       \textbf{\Ours-VAR-$d$30} & AR & 2.0B & 10 & 2.0 & \textbf{1.59} & 324.3 \\
       \bottomrule
    \end{tabular}
    }
    \centering
    \begin{minipage}{.88\linewidth}
       \vspace{.2em}
       \scriptsize
       TFLOPs quantify the total computational cost of generating a single image.
       $^\dagger$: uses rejection sampling.
    \end{minipage}
	\vspace{-1em}
\end{table*}

\begin{table}[!t]
    \caption{
       {Class-conditional Image Generation on ImageNet 512$\times$512\label{tab:fid_imnet512}}
    }
    \vspace{-.5em}
    \centering
    \tablestyle{5pt}{1}
    \begin{tabular}{y{72}x{29}x{19}x{38}x{41}}
       \toprule
       Method & Params & Steps & TFLOPs$\downarrow$ & FID-50K$\downarrow$  \\
       \midrule
       VQGAN~\cite{esser2021taming}                & 227M & 1024    & -     & 26.52   \\
       ADM-G~\cite{dhariwal2021diffusion}               & 559M & 250     & 579.0 & 7.72  \\
        U-ViT-H~\cite{bao2022all} & 501M & 50 & 14.6 & 4.05 \\
       DiT-XL~\cite{peebles2023scalable} & 675M & 250 & 60.0 & 3.04 \\
        RIN~\cite{jabri2022scalable} & 320M & 1000 & 415.0 & 3.95 \\
        Simple-Diffusion~\cite{hoogeboom2023simple} & 2B & 512 & - & 3.02 \\
        VDM++~\cite{kingma2023understanding} & 2.4B & 256 & - & 2.65 \\
        EDM2-XS~\cite{karras2024analyzing} & 125M & 63 & 5.7 & 2.91 \\
        DiffiT~\cite{hatamizadeh2024diffit} & 561M & 250 & - & 2.67 \\
        Large-DiT-3B-G~\cite{gao2024lumina} & 3B & 250 & - & 2.52 \\
        MaskDiT~\cite{zheng2023fast} & 736M & 79 & - & 2.50 \\
        DyDiT-XL~\cite{zhao2024dynamic} & 678M & 100 & 75.1 & 3.61 \\
       \midrule
       MaskGIT~\cite{chang2022maskgit} & 227M & 12 & 3.3 & 7.32 \\
       {MaskGIT$^{\dagger}$~\cite{chang2022maskgit}} & 227M & 12 & 13.1 & 4.46 \\
       Token-Critic~\cite{lezama2022improved} & 422M & 36 & 7.6 & 6.80  \\
       {Token-Critic$^{\dagger}$~\cite{lezama2022improved}} & 422M & 36 & 34.8 & 4.03  \\

       \midrule
       \multirow{2}{*}{\textbf{\Ours-MaskGIT-L}} & 232M & 16 & \textbf{2.0} & 2.82 \\
       & 232M & 32 & 3.6 & \textbf{2.46} \\
       \bottomrule
    \end{tabular}
    \begin{minipage}{.99\linewidth}
       \vspace{.2em}
       \scriptsize
       $^\dagger$: uses rejection sampling.
    \end{minipage}
	\vspace{-1em}
\end{table}

\begin{figure}[t!]
    \centering
    \includegraphics[width=\linewidth]{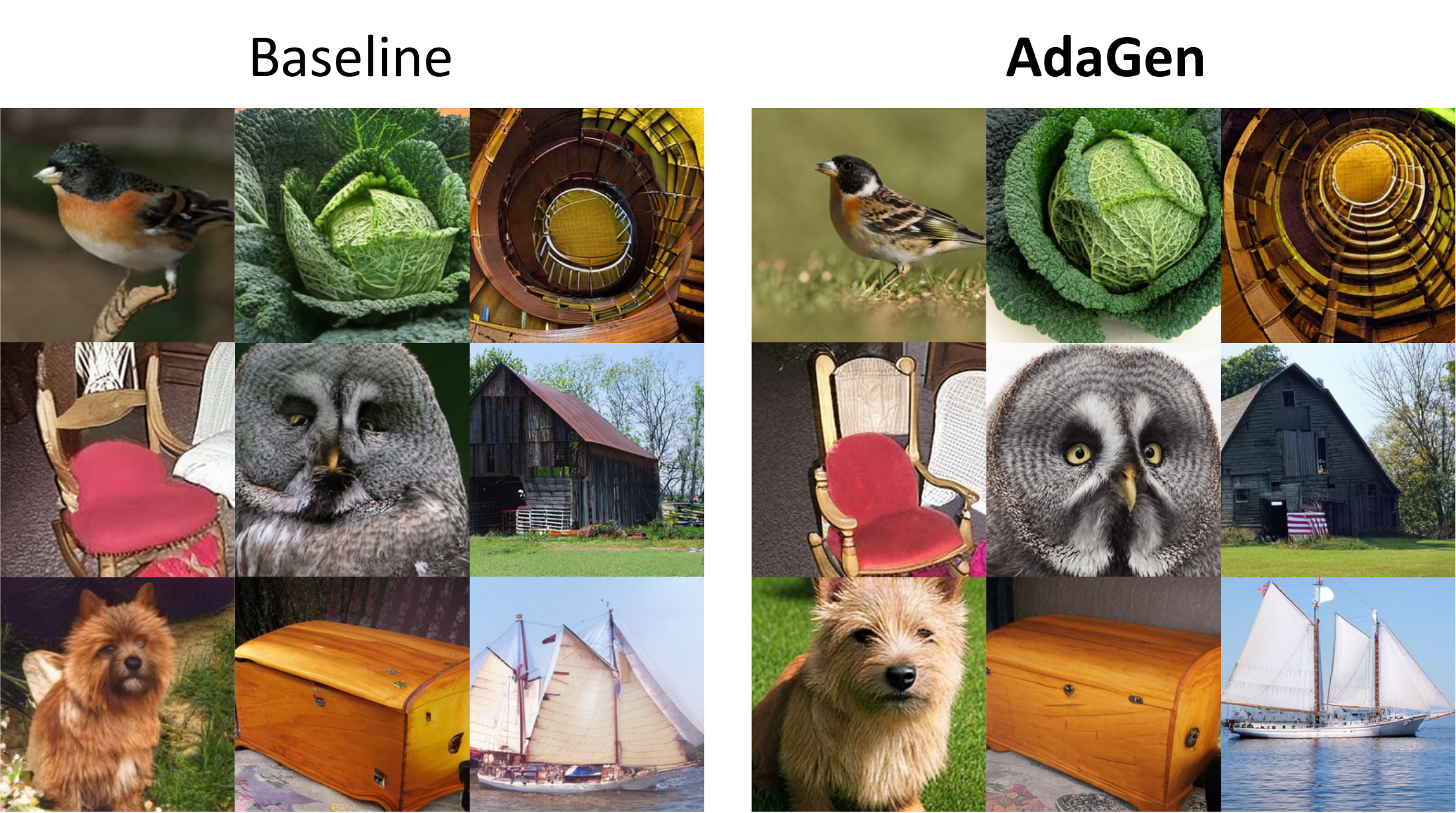}
    \vspace{-1.5em}
    \caption{
       \textbf{Effectiveness of \Ours on class-conditional image generation}.
	   We visualize the generated samples of VAR-$d$30 without and with \Ours.
    }
    \label{fig:compare}
	\vspace{-1em}
\end{figure}

\paragraph{Qualitative Results.}
\begin{figure*}[t!]
    \centering
    \includegraphics[width=\linewidth]{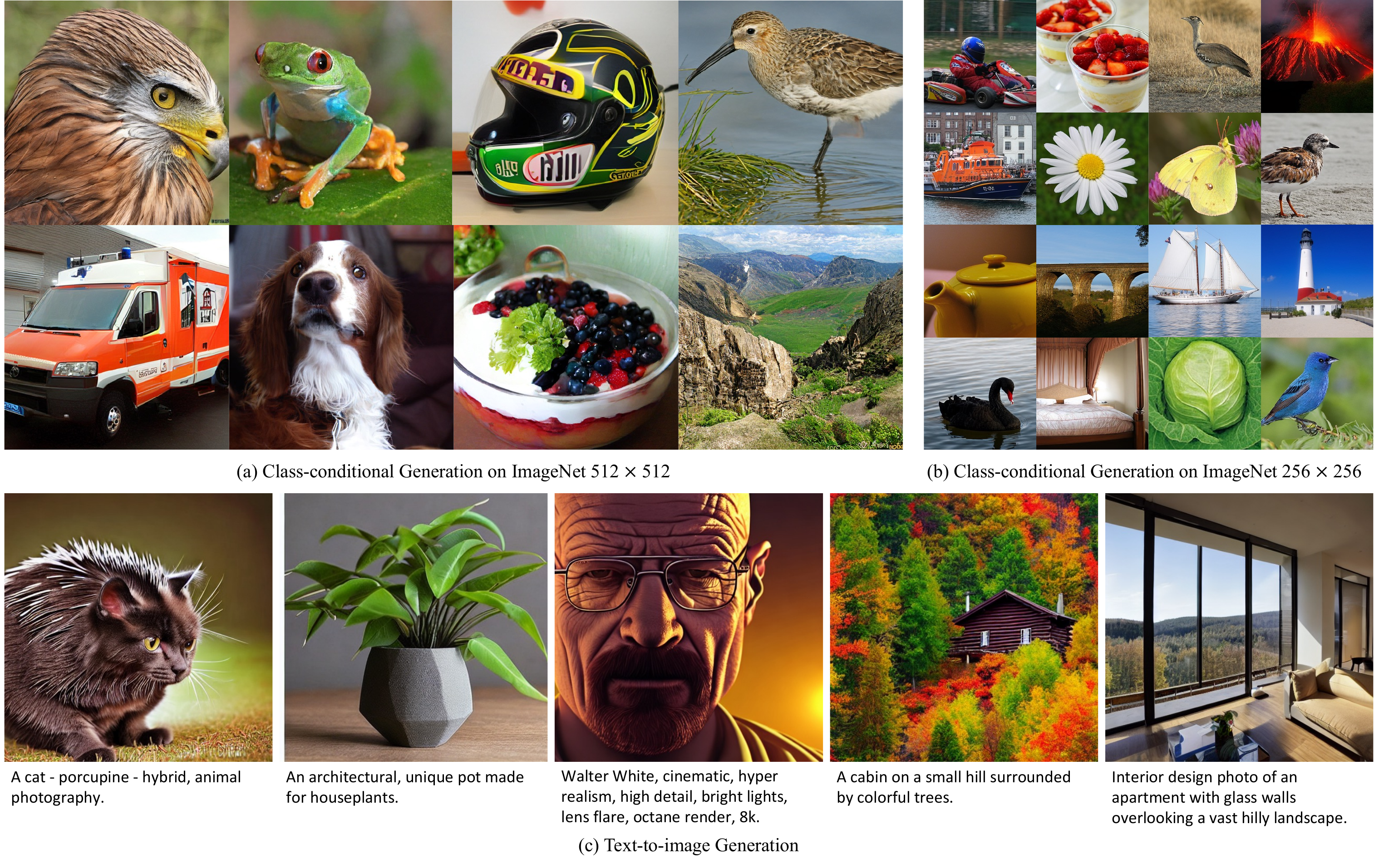}
    \vspace{-2.5em}
    \caption{\textbf{Qualitative results of \Ours}.
    The corresponding models are (a) \Ours-MaskGIT-L ($T=32$), (b) \Ours-VAR-$d$30 ($T=10$) and (c) \Ours-Stable-Diffusion ($T=32$), respectively.
    }
	\label{fig:quality_results}
	\vspace{-1.5em}
\end{figure*}
Figure~\ref{fig:quality_results}a and ~\ref{fig:quality_results}b present qualitative results on ImageNet at 512$\times$512 and 256$\times$256 resolutions, respectively. Our method generates high-fidelity images across diverse categories. In Figure~\ref{fig:compare}, we provide a direct visual comparison between the VAR-$d$30 models with and without our adaptive policy. The results clearly illustrate that \Ours consistently enhances visual quality, reduces artifacts, and preserves fine details across various image categories.
We provide more qualitative results in Appendix~\ref{sec:app_additional_qual}.

\subsection{Text-to-image Generation}
\label{sec:t2i_exp}
\paragraph{Text-to-image Generation on MS-COCO.}
As shown in Table~\ref{tab:fid_mscoco}, \Ours-MaskGIT-S achieves a compelling performance on MS-COCO, reaching an FID score of 4.92 with 16 sampling steps. This marks a clear improvement over the MaskGIT-S baseline (5.78 FID) under the same computational cost. Moreover, our 8-step variant attains an FID of 5.29 with only 0.3 TFLOPs, outperforming both the 16-step MaskGIT-S baseline at 5.78 FID and 0.6 TFLOPs, and the 50-step U-ViT model, which achieves 5.48 FID with a cost of 2.2 TFLOPs.

\begin{table}[t!]
    \centering
    \caption{
       {Text-to-image Generation on MS-COCO}
       \label{tab:fid_mscoco}
    }
    \vspace{-.5em}
    \tablestyle{3pt}{1}
    \begin{tabular}
    {y{72}x{29}x{19}x{38}x{41}}
       \toprule
       Method & Params &  Steps & TFLOPs$\downarrow$ & FID-30K$\downarrow$\\
       \midrule
       VQ-Diffusion~\cite{gu2022vector} & 370M & 100 & 206.7 & 13.86 \\
       Frido~\cite{fan2023frido} &512M & 200 & - & 8.97 \\
       Make-A-Scene~\cite{gafni2022make} & 4B & 1536 & - & 7.55 \\
       U-Net~\cite{bao2022all} &53M & 50 & - & 7.32 \\
       MaskGIT-S~\cite{chang2022maskgit} & 45M & 16 & 0.6 & 5.78 \\
       \multirow{2}{*}{U-ViT~\cite{bao2022all}} & \multirow{2}{*}{58M} & 16 & 0.9 & 5.91 \\
       &  & 50 & 2.2 & 5.48 \\
       \midrule
       \multirow{2}{*}{\textbf{\Ours-MaskGIT-S}} & \multirow{2}{*}{57M} & 8 & \textbf{0.3} & 5.29 \\
       &  & 16 & 0.6 & \textbf{4.92} \\
       \bottomrule
    \end{tabular}
	\vspace{-1em}
\end{table}
\paragraph{Text-to-image Generation on CC3M.}
As shown in Table~\ref{tab:fid_cc3m}, we apply \Ours to the larger text-to-image model Muse~\cite{chang2023muse} and evaluate its performance on the CC3M dataset. Our method demonstrates better efficiency and performance, achieving an FID-30K score of 6.66 with only 8 inference steps, already surpassing the Muse baseline at 16 steps while consuming roughly half the computation. With 16 steps, \Ours-Muse further improves to an FID of 6.36, clearly outperforming the corresponding baseline.

\begin{table}[t!]
    \begin{center}
    \caption{
       {Text-to-image Generation on CC3M}
	   \label{tab:fid_cc3m}
    }
    \vspace{-.5em}
    \tablestyle{3pt}{1}
    \begin{tabular}{y{83}x{29}x{19}x{38}x{41}}
       \toprule
       Method                    & Params             & Steps & TFLOPs$\downarrow$ & FID-30K$\downarrow$   \\
       \midrule
        VQGAN~\cite{esser2021taming}                     & 600M                  & 256                &   12.6    & 28.86 \\
        ImageBART~\cite{esser2021imagebart} & 2.8B & 256 & 7.4 & 22.61 \\
       LDM-4~\cite{rombach2022high}                     & 645M                  & 50                  &   -    & 17.01 \\
       Draft-and-revise~\cite{lee2022draft}          & 654M                  & 72                  &   -    & 9.65  \\
       \multirow{2}{*}{Muse~\cite{chang2023muse}} & \multirow{2}{*}{500M} & 8                  &    \textbf{2.8}    & 7.67  \\
        &  & 16                  &    5.4    & 7.01  \\
       \midrule
       \multirow{2}{*}{\textbf{\Ours-Muse}} & \multirow{2}{*}{512M} & 8 & \textbf{2.8} & 6.66 \\
       &  & 16 & 5.4 & \textbf{6.36} \\
       \bottomrule
    \end{tabular}
    \end{center}
	\vspace{-1.3em}
\end{table}

\paragraph{Large-scale Text-to-image Generation on LAION-5B.}
To evaluate the effectiveness of our approach in large-scale settings, we integrated \Ours into Stable Diffusion and conducted zero-shot evaluation on MS-COCO, following~\cite{saharia2022photorealistic,ramesh2022hierarchical}. As reported in Table~\ref{tab:fid_mscoco_zero_shot}, \Ours-Stable Diffusion achieves 8.14 FID using 32 inference steps, outperforming the original model's 9.03 FID under the same computational budget. It also surpasses the 50-step Stable Diffusion baseline at 8.59 FID while requiring roughly $1.5\times$ less computation. These results demonstrate that our method remains effective when deployed in large-scale generative systems.

\begin{table}[t!]
    \centering
    \caption{
       {Zero-shot Text-to-image Generation on MSCOCO}\label{tab:fid_mscoco_zero_shot}
    }
    \vspace{-.5em}
    \tablestyle{3pt}{1}
    \begin{tabular}{y{99}cccc}
       \toprule
       Method                    & Params             & Steps & TFLOPs$\downarrow$ & FID-30K$\downarrow$   \\
       \midrule
        CogView~\cite{ding2021cogview} & 4B & 1088 & 182.2 & 27.1 \\
        LAFITE~\cite{zhou2022lafite} & 0.2B & - & - & 26.94 \\
       CogView 2~\cite{ding2022cogview2} & 6B & 1088 & 117.0 & 24.00 \\
       DALL-E~\cite{esser2021taming}                     & 12B                  & 256                &   309.8    & 17.89 \\
        GLIDE~\cite{nichol2021glide} & 5B  & 250 & 77.0  & 12.24 \\
        Make-A-Scene~\cite{gafni2022make} & 4B & 1536 & - & 11.84 \\
       Corgi~\cite{zhou2023shifted} & 5.7B & 64 & 43.3 & 10.88 \\
       DALL-E 2~\cite{ramesh2022hierarchical} & 6.5B & 250 & - & 10.39 \\
        \multirow{2}{*}{Stable Diffusion~\cite{rombach2022high}} & \multirow{2}{*}{1.4B} & 32 & \textbf{22.9} & 9.03 \\
       & & 50 & 35.1 & 8.59 \\
       Simple Diffusion~\cite{hoogeboom2023simple} & 2B & 512 & 151.3 & 8.30 \\
       \midrule
       \textbf{\Ours-Stable Diffusion} & 1.4B & 32 & \textbf{22.9} & \textbf{8.14} \\
       \bottomrule
    \end{tabular}
\end{table}

\begin{figure}[t!]
    \centering
    \includegraphics[width=.9\linewidth]{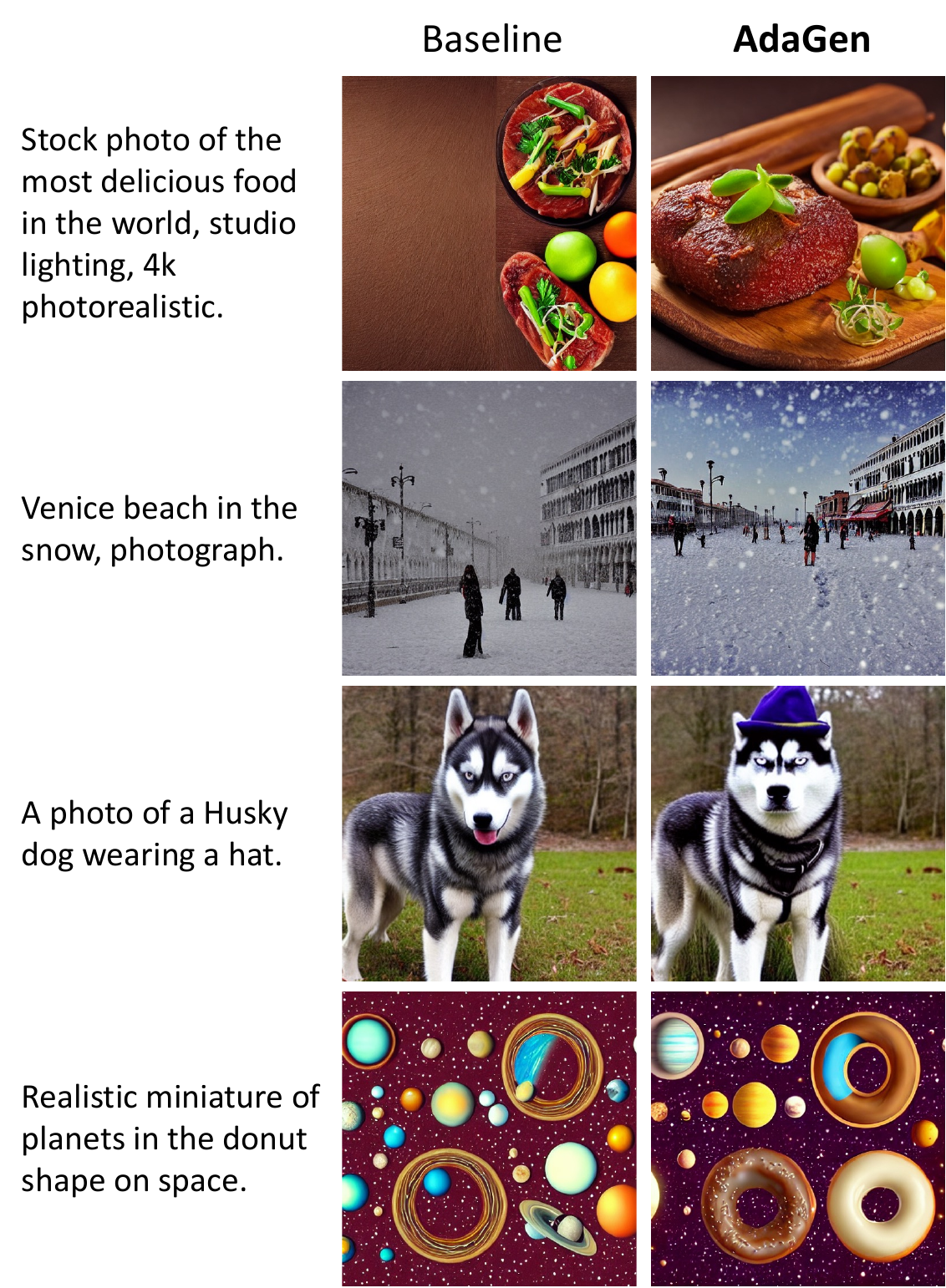}
	\vspace{-.5em}
    \caption{
       \textbf{Effectiveness of \Ours on text-to-image generation}.
	   We visualize the generated samples of Stable Diffusion without and with \Ours.
    }
    \label{fig:sd_compare}
	\vspace{-.5em}
\end{figure}

\paragraph{Qualitative Results.}
Figure~\ref{fig:quality_results}c showcases \Ours-Stable Diffusion's ability to generate high-fidelity, prompt-aligned images.
Figure~\ref{fig:sd_compare} contrasts the image generation quality of the standard Stable Diffusion baseline against our method. As demonstrated in the first two rows, AdaGen produces images with superior aesthetic quality. Furthermore, the last two rows highlight its improved image-text alignment. For the prompt ``A photo of a Husky dog wearing a hat'', the baseline fails to generate the hat, an attribute that AdaGen successfully includes. Similarly, when prompted to create a ``Realistic miniature of planets in the donut shape on space'', the baseline model struggles with the specified donut shapes, whereas AdaGen accurately renders the planets as donuts with convincing textures and appearance.
We provide more qualitative results in Appendix~\ref{sec:app_additional_qual}.

\subsection{Inference-time Refinement}
\label{sec:inference_refinement_exp}

As introduced in Section~\ref{sec:inference_refinement}, we repurpose the adversarial reward model $r_{\bm{\psi}}$ as a perceptual evaluator for repeated sampling, while the value network $V_{\bm{\phi}}$ further enables lookahead refinement in models with stochastic transitions. We evaluate these strategies on MaskGIT-L with $T = 32$ and DiT-XL with $T = 16$.
As shown in Table~\ref{tab:inference_refinement}, repeated sampling consistently improves performance for both models. For MaskGIT-L, where transitions are stochastic, lookahead refinement (with $K = 2$ branches, which we find empirically sufficient) yields additional gains, reducing FID to 1.94.

We further study the impact of the number of repeated sampling trials \( M \), with results in Table~\ref{tab:inference_refinement2}. Increasing \( M \) leads to steady improvements—for example, in MaskGIT-L, FID drops from 2.28 to 1.81 with three sampling iterations (lookahead sampling is also enabled during generation). These results confirm the general effectiveness of our inference-time refinement. We also provide qualitative results in Appendix~\ref{sec:app_additional_qual}.

\begin{table}[t]
    \centering
    \caption{
       {Inference-time Refinement of \Ours}
       \label{tab:inference_refinement}
    }
    \vspace{-1em}
    \begin{center}\vspace{-.4em}
    \tablestyle{1.5mm}{1}
    \begin{tabular}{cccc}
       \toprule
       & Repeated Sampling? & Lookahead Sampling? & FID-50K$\downarrow$ \\
       \midrule
       \multirow{3}{*}{MaskGIT-L} & {\xmark} & {\xmark} & {2.28} \\
          & {\cmark} & {\xmark} & {2.07} \\
          & {\cmark} & {\cmark} & {\textbf{1.94}} \\
       \midrule
       \multirow{3}{*}{DiT-XL} & {\xmark} & {\xmark} & {2.19} \\
          & \cmark & \xmark & \textbf{2.06} \\
          & \gc{\cmark} & \gc{\cmark} & \gc{-} \\
       \bottomrule
    \end{tabular}\vspace{-.4em}
    \end{center}
    \begin{minipage}{.96\columnwidth}
       \scriptsize
       We apply inference-time refinement to MaskGIT-L ($T=32$) and DiT-XL ($T=16$).
       For DiT-XL, since the sampling process adopts an ODE solver~\cite{lu2022dpmp} and thus the state transition is deterministic, we only apply repeated sampling.
       Here, repeated sampling is executed once, and Table~\ref{tab:inference_refinement2} shows the results with more repeated sampling trials.
    \end{minipage}
	\vspace{-1em}
\end{table}

\begin{table}[t]
    \centering
    \caption{
       {Inference-time Refinement of \Ours with Different Numbers of Repeated Sampling Trials}
       \label{tab:inference_refinement2}
    }
    \vspace{-.8em}
    \begin{center}\vspace{-.4em}
    \tablestyle{2mm}{.95}
    \begin{tabular}{cccccccc}
       \toprule
       \# Repeat Trials & 0  & 1 & 2 & 3 & 4 & 5 & 6\\
       \midrule
       MaskGIT-L & 2.28 & 1.94 & 1.85 & \textbf{1.81} & 1.84 & 1.89 & 1.90 \\
       DiT-XL & 2.19 & 2.06 & 2.02 & 1.99 & 1.93 & \textbf{1.91} & 1.92 \\
       \bottomrule
    \end{tabular}
    \end{center}
	\vspace{-1.2em}
\end{table}

\subsection{Flexibly Trading Off Fidelity and Diversity.}
\label{sec:tradeoff_exp}
\begin{figure*}[t!]
    \centering
    \includegraphics[width=.95\linewidth]{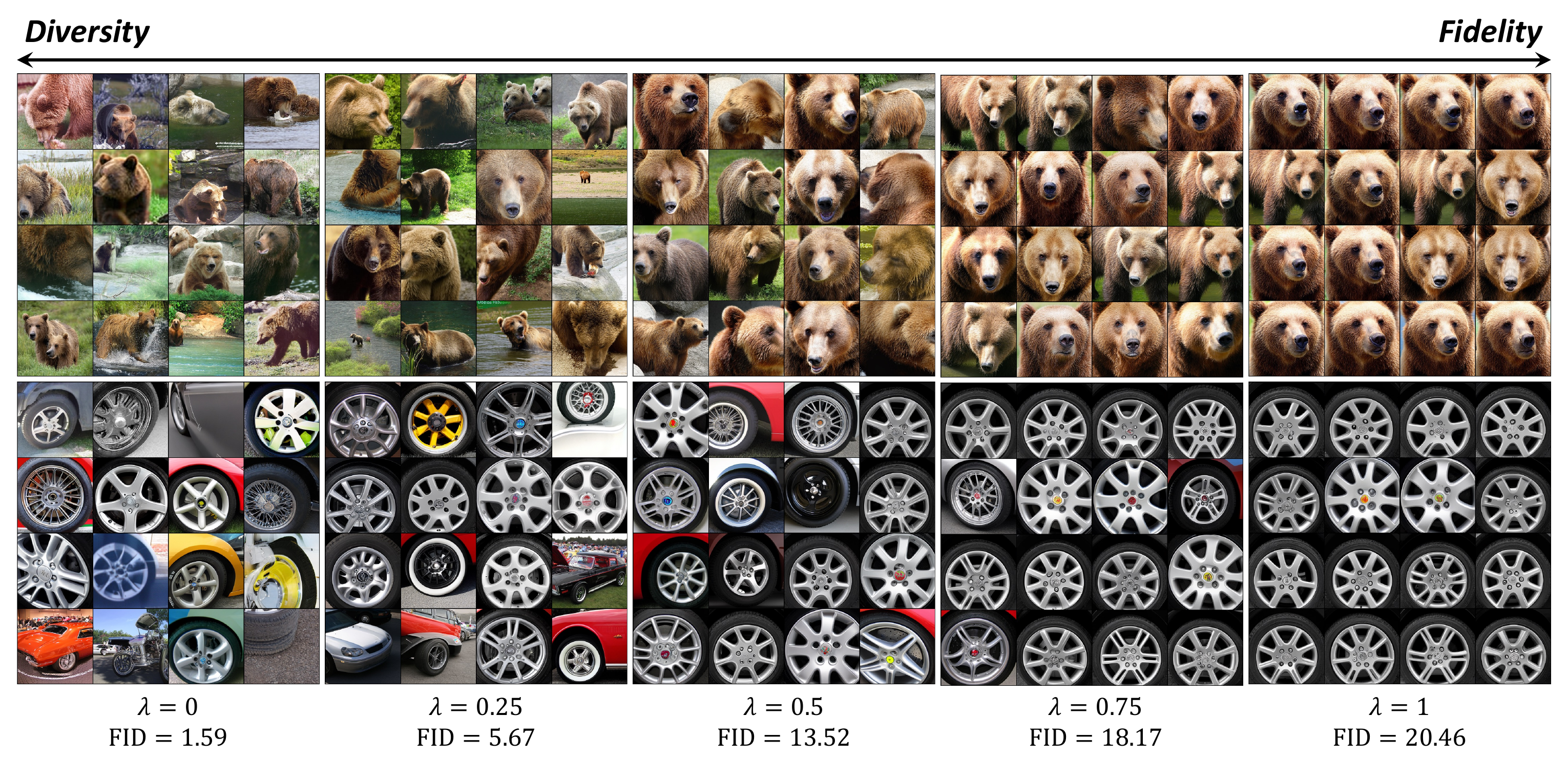}
	\vspace{-1em}
    \caption{
       \textbf{The fidelity-diversity trade-off of AdaGen}. We visualize 4$\times$4 samples from two ImageNet classes (brown bears and car wheels) across different $\lambda$ values using AdaGen-VAR-$d$30. Each setting reports FID scores measuring the fidelity-diversity balance. Higher $\lambda$ values prioritize fidelity over diversity.
    }
    \label{fig:tradeoff_fid_div}
	\vspace{-1em}
\end{figure*}
Section~\ref{sec:tradeoff} introduced a blending policy with parameter \( \lambda \) to control the fidelity-diversity trade-off. We evaluate its effectiveness on ImageNet 256\(\times\)256 by varying \( \lambda \in \{0, 0.25, 0.5, 0.75, 1.0\} \), where larger values favor fidelity over diversity. Figure~\ref{fig:tradeoff_fid_div} illustrates this trade-off on two ImageNet classes with 4\(\times\)4 visual grids, along with the corresponding FID scores which capture the fidelity-diversity balance.
At \( \lambda = 0 \), outputs are highly diverse but may occasionally contain visual artifacts. Increasing \( \lambda \) to 0.25 and 0.5 eliminates lower-quality samples while maintaining reasonable diversity. At \( \lambda = 0.75 \), the model yields consistently realistic and high-quality samples. At \( \lambda = 1.0 \), fidelity peaks with visually appealing but highly converged outputs and reduced diversity. This controllable trade-off allows practitioners to tailor the balance to their specific application needs.

\subsection{Discussions}
\label{sec:analysis}
This section presents more analyses of \Ours.
If not mentioned otherwise, we use \Ours-MaskGIT-S ($T=4$) by default.

\begin{figure}[t!]\centering
    \includegraphics[width=\linewidth]{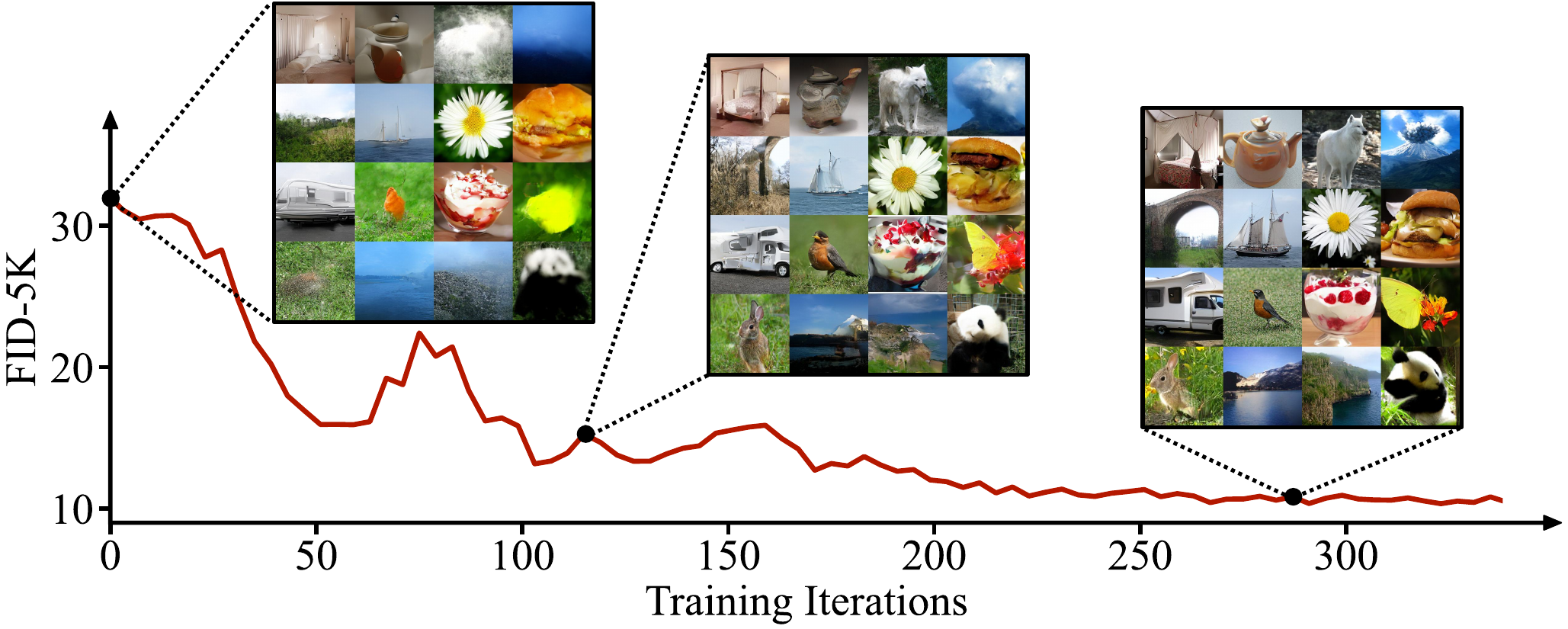}
    \caption{
       \textbf{Optimization curve of \Ours with intermediate samples visualized.}
	   FID-5K scores are reported for efficiency.
    }
    \label{fig:optimization_curve}
	\vspace{-1em}
\end{figure}

\begin{table}[t]
    \centering
    \caption{
       {Effectiveness of \Ours}
       \label{tab:effectiveness}
    }
    \vspace{-.5em}
    \tablestyle{6pt}{1.15}
    \begin{tabular}{x{50}x{50}|x{50}x{50}}
       \toprule
       \multicolumn{2}{c|}{Generation Policy} & \multicolumn{2}{c}{FID-50K$\downarrow$}\\
       \emph{Learnable?} & \emph{Adaptive?} & MaskGIT-S & DiT-XL \\
       \hline
       \gc{\xmark} & \gc{\xmark} & \gc{7.65} & \gc{9.71} \\
       \cmark     & \xmark & 5.40\fontsize{5.5pt}{0.1em}\selectfont~\hl{(-2.25)} & 6.03\fontsize{5.5pt}{0.1em}\selectfont~\hl{(-3.68)} \\
       \cmark & \cmark & \textbf{4.54}\fontsize{5.5pt}{0.1em}\selectfont~\hl{(-3.11)} & \textbf{5.31}\fontsize{5.5pt}{0.1em}\selectfont~\hl{(-4.40)} \\
       \bottomrule
    \end{tabular}
    \begin{minipage}{.96\columnwidth}
       \vspace{.3em}
       \scriptsize
       We use MaskGIT-S and DiT-XL ($T=4$) on ImageNet 256$\times$256.
    \end{minipage}
\end{table}

\paragraph{Effectiveness of \Ours.}
\label{sec:effectiveness}
In Table~\ref{tab:effectiveness}, we evaluate the effectiveness of \Ours using MaskGIT-S and DiT-XL with $T=4$. Introducing learnability yields substantial improvements: MaskGIT-S achieves a 2.25 FID reduction from 7.65 to 5.40, a 30\% relative improvement, while DiT-XL obtains a 3.68 reduction from 9.71 to 6.03, a 38\% gain. These results suggest the sub-optimality of the hand-crafted generation policies from prior works~\cite{chang2022maskgit,peebles2023scalable}.
Making the policy both learnable and adaptive further improves performance. MaskGIT-S reaches an FID of 4.54 with an additional 16\% improvement, while DiT-XL achieves 5.31 with a 12\% gain. These consistent improvements demonstrate the effectiveness of learnable, adaptive policies.

Figure~\ref{fig:optimization_curve} visualizes generation results during policy training. Initially, the policy produced very blurry images with high FID scores. As training progresses, the policy gradually refines itself, yielding lower FID scores and improved image quality. Finally, the policy converges to a robust generation strategy that consistently produces quality images with low FID scores, demonstrating the effectiveness of \Ours.

\paragraph{Visualizing the Learned Adaptive Policy.}
\begin{figure}[t!]\centering
    \includegraphics[width=\linewidth]{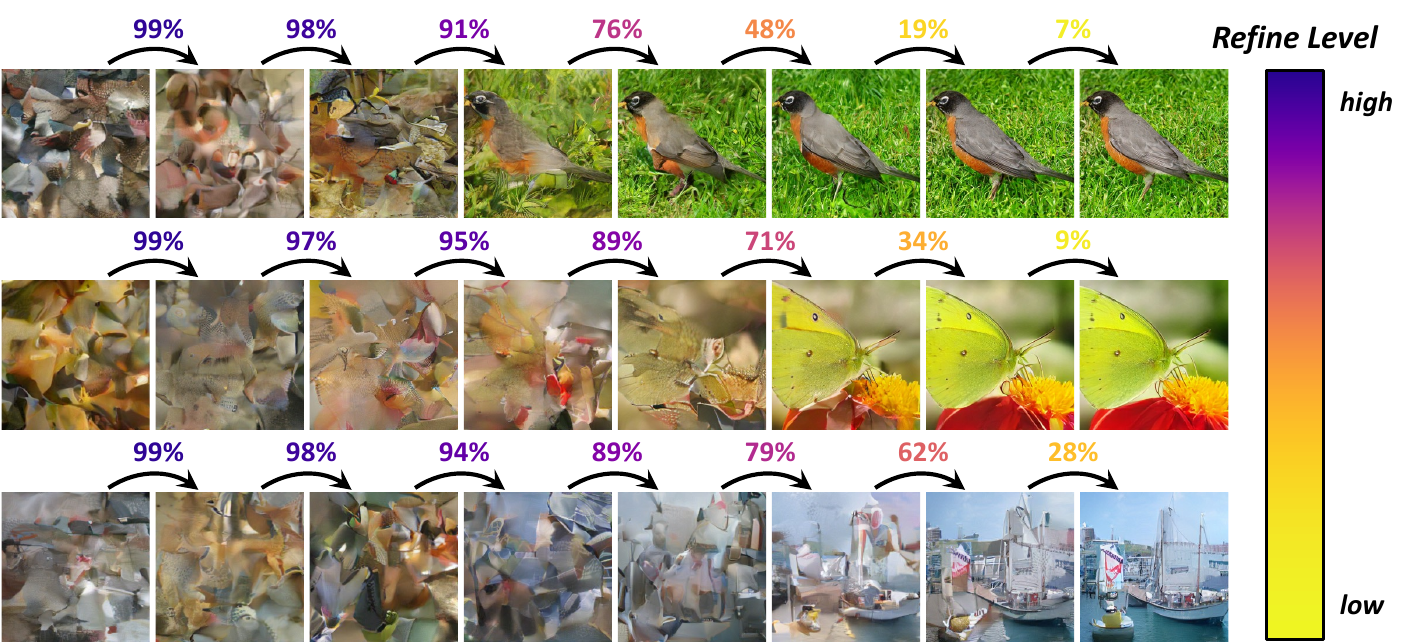}
	\vspace{-1.7em}
    \caption{
       \textbf{Visualizing the learned policies of \Ours.}
       We use \Ours-MaskGIT-L ($T=8$) on ImageNet 256$\times$256 as an example.
       The mask ratio $m_t$ (refine level), which controls the proportion of least-confident tokens to be refined at each step in MaskGIT (see~\cite{chang2022maskgit} for more details), is visualized as an example. The policy network adaptively reduces $m_t$ for only minor refinements when the sample already reaches a decent quality; otherwise, it keeps adopting relatively higher $m_t$ for more adjustments.
    }
    \label{fig:vis_policy}
	\vspace{-1em}
\end{figure}
Figure~\ref{fig:vis_policy} illustrates our learned policies using \Ours-MaskGIT-L with $T=8$ as an example. We visualize the adaptive policy with mask ratio $m_t$, which controls the proportion of least-confident tokens to be refined at each step. For samples with simple structures (first row), the refine level quickly drops once reasonable quality is achieved, limiting refinements. For harder structures (second row), the policy maintains higher refine levels for more adjustments. When the structure of the sample becomes complex (third row), the policy consistently applies relatively high refine levels to make more refinements until the end. These visualizations provide deeper insights into the adaptive mechanism of \Ours.

\setcounter{table}{14}
\begin{table*}[t]
    \centering
    \caption{
    Ablation Studies on Model Architecture and Components of \Ours.
    Our Default Setting is Marked in \colorbox{defaultcolor}{Gray}.
    }
    \vspace{-2.1em}
    \label{tab:ablations}
    \subfloat[
        \textbf{Discriminator architecture.}
        The transformer-based architecture~\cite{sauer2023stylegan} significantly outperforms the baseline and Convolution-based architecture~\cite{Karras2019AnalyzingAI}.
        \label{tab:arch_comparison}
    ]{
        \begin{minipage}{0.32\linewidth}{
            \begin{center}
                \tablestyle{3pt}{1.15}
                \begin{tabular}{x{65}x{30}x{40}}
                \toprule
                    Architecture & \# Params & FID-50K$\downarrow$ \\
                \midrule
                    Baseline & / & 7.65 \\
                    \default{Transformer-based} & \default{32.0M} & \default{\textbf{4.54}} \\
                    Conv-based & 26.9M & 5.86 \\
                \bottomrule
                \end{tabular}
            \end{center}
        }\end{minipage}
    }
\hspace{1em}
    \subfloat[
    \textbf{Output activation.}
    Mapping the policy network's output to the valid range via activation functions (see Appendix~\ref{sec:imp_details}) is more effective than clamping.
    \label{tab:output_activation}
    ]{
        \begin{minipage}{0.335\linewidth}{\begin{center}
             \tablestyle{3pt}{1.15}
             \begin{tabular}{x{60}x{40}}
                \toprule
                 Output Activation & FID-50K$\downarrow$ \\
                \midrule
                 \default{Activation} & \default{\textbf{4.54}} \\
                 Simple Clamping & 6.29 \\
                \bottomrule
                \multicolumn{2}{c}{~}\\
             \end{tabular}
        \end{center}}\end{minipage}
    }
\hspace{1em}
    \subfloat[
        \textbf{Input type.}
       Using off-the-shelf features from the generative model as the policy network's input is more effective.
        \label{tab:input_type}
    ]{
        \centering
        \begin{minipage}{0.24\linewidth}{\begin{center}
            \tablestyle{3pt}{1.15}
            \begin{tabular}{x{50}x{40}}
             \toprule
                Input Type & FID-50K$\downarrow$ \\
             \midrule
                \default{Features} & \default{\textbf{4.54}} \\
                Raw Input & 6.55 \\
             \bottomrule
             \multicolumn{2}{c}{~}\\
            \end{tabular}
        \end{center}}\end{minipage}
    }
    \hspace{.8em} \\
    \vspace{-1em}
\subfloat[
    \textbf{Policy network architecture.}
    The combination of a convolutional layer and an MLP block already achieves a satisfactory performance with low parameter count and computational cost.
    \label{tab:arch_design}
]{
    \begin{minipage}{0.45\linewidth}{\begin{center}
        \tablestyle{3pt}{1.15}
        \begin{tabular}{{z{40}x{5}y{40}x{25}x{25}x{40}}}
          \toprule
            \multicolumn{3}{c}{Architecture} & Params$\downarrow$ & FLOPs$\downarrow$ & FID-50K$\downarrow$ \\
          \midrule
            \default{Conv} & \default{+} & \default{MLP} & \default{{12.0M}} & \default{{0.016G}} & \default{4.54} \\
            Trans Block & + & MLP & 16.7M & 0.891G & 4.72 \\
          Conv & + & Linear & 6.3M & 0.006G & 5.03 \\
            Conv & + & Heavy MLP & 24.6M & 0.029G & \textbf{4.49} \\
          \bottomrule
        \end{tabular}
    \end{center}}\end{minipage}
}
\hspace{3em}
\subfloat[
    \textbf{Generation step conditioning.}
    Incorporating the current generation step $t$ into the policy network is critical for effective policy decisions.
    \label{tab:step_cond}
]{
    \begin{minipage}{0.33\linewidth}{\begin{center}
        \tablestyle{3pt}{1.15}
        \begin{tabular}{cc}
          \toprule
            Step Cond. & FID-50K$\downarrow$ \\
          \midrule
            \default{\cmark} & \default{\textbf{4.54}} \\
          \xmark & 6.13 \\
          \bottomrule
          \multicolumn{2}{c}{~}\\
          \multicolumn{2}{c}{~}\\
        \end{tabular}
    \end{center}}\end{minipage}
}
    \vspace{-1.7em}
\end{table*}

\paragraph{Computational Overhead of \Ours.}
Table~\ref{tab:computational_overhead} provides a detailed breakdown of inference cost, comparing our policy network's computational requirements with various generative models. Results show minimal overhead of only $0.07\%$-$0.40\%$ added to the original generator's computational cost. This demonstrates the practicality of achieving notable performance gains with negligible computational overhead.

\setcounter{table}{11}
\begin{table}[t]
    \centering
    \caption{Computational Overhead of \Ours}
    \label{tab:computational_overhead}
    \vspace{-.6em}
    \tablestyle{1mm}{1.2}
    \begin{tabular}{x{40}x{10}|x{40}x{60}x{70}}
       \toprule
       \multirow{2}{*}{Generator} & \multirow{2}{*}{$T$} & \multicolumn{2}{c}{Inference Cost (TFLOPs)} & Proportion \\ 
       & & Generator & Policy Net. & (Policy / Generator) \\
       \hline
        MaskGIT-L & 16 & 1.7 & 0.002 &  0.11\% \\
        DiT-XL & 16 & 4.1 & 0.003 & 0.07\% \\
        SiT-XL & 16 & 4.1 & 0.003 & 0.07\% \\
        VAR-$d$16 & 10 & 0.5 & 0.002 & 0.40\% \\
        VAR-$d$30 & 10 & 2.0 & 0.003 & 0.15\% \\
        \bottomrule
    \end{tabular}
    \begin{minipage}{.98\columnwidth}
       \vspace{.3em}
       \scriptsize
       We report the inference cost (in TFLOPs) for various generative models and our policy networks on ImageNet. The proportion column indicates the relative computational overhead introduced by our policy network compared to the base generator model.
    \end{minipage}
	\vspace{-.9em}
\end{table}

\paragraph{Sensitivity Test on PPO Hyperparameters.}
Our policy network is optimized using the PPO algorithm~\cite{schulman2017proximal}, which introduces several hyperparameters defined in Eqs.~\eqref{eq:policy_to_agent} and~\eqref{eq:ppo}. Table~\ref{tab:ppo_hyp} shows that our method is able to achieve superior performance across a variety of hyperparameter choices, demonstrating its robustness.

\begin{table}[t]
  \caption{
    {Sensitivity Analysis of PPO Hyperparameters.}
    Default Setting is Marked in \colorbox{defaultcolor}{Gray}.
    \Ours Consistently Outperforms the Baseline \textcolor{deemph}{(Light Gray)} across Various Hyperparameter Choices.
  }
  \vspace{-1.5em}
  \label{tab:ppo_hyp}

  \noindent
  \resizebox{0.75\columnwidth}{!}{%
    \begin{minipage}{\columnwidth}
      \begin{tabular}{@{}c@{\hspace{0.05\columnwidth}}c@{\hspace{0.05\columnwidth}}c@{}}
        \subfloat[
          Effects of varying $\epsilon$.
        ]{%
          \begin{tabular}{x{30}x{45}}
          \toprule
            $\epsilon$ & FID-50K$\downarrow$ \\
          \midrule
            \gc{-}      & \gc{7.65} \\
            0.05        & 5.01     \\
            0.1         & 4.86     \\
            \default{0.2} & \default{4.54} \\
            0.3         & \textbf{4.45} \\
            0.4         & 4.68     \\
          \bottomrule
          \end{tabular}
        }
        &
        \subfloat[
          Effects of varying $c$.
        ]{%
          \begin{tabular}{x{30}x{45}}
          \toprule
            $c$   & FID-50K$\downarrow$ \\
          \midrule
          \gc{-} & \gc{7.65} \\
            0.2   & 4.63               \\
            0.4   & 4.59               \\
            \default{0.5} & \default{\textbf{4.54}} \\
            0.6   & 4.66               \\
            0.8   & 4.73               \\
          \bottomrule
          \end{tabular}
        }
        &
        \subfloat[
          Effects of varying $\sigma$.
        ]{%
          \begin{tabular}{x{30}x{45}}
          \toprule
            $\sigma$ & FID-50K$\downarrow$ \\
          \midrule
          \gc{-} & \gc{7.65} \\
            0.1      & 5.31               \\
            0.3      & 4.72               \\
            \default{0.6} & \default{\textbf{4.54}} \\
            0.8      & 4.66               \\
            1.0      & 4.73               \\
          \bottomrule
          \end{tabular}
        }
      \end{tabular}
    \end{minipage}%
  }
  \vspace{-1em}
\end{table}

\paragraph{Effect of Action Smoothing Weight.}
As discussed in Section~\ref{sec:action_stablize}, we employ action smoothing to stabilize exploration, introducing smoothing hyperparameter $\beta$. 
Table~\ref{tab:abl_momentum} reports FID-50K scores for a variety of $\beta$ values, with $\beta$ of 0 representing no smoothing serving as the baseline. Across all $\beta$ values, our action smoothing consistently outperforms the baseline and achieves the lowest FID of 3.36 with $\beta$ of 0.8. As a result, we set $\beta$ to 0.8 in our experiments.

\begin{table}[t]
    \caption{Effect of Action Smoothing Weight $\beta$}
    \vspace{-1.5em}
\begin{center}
    \tablestyle{3mm}{1.2}
    \begin{tabular}{c|ccc>{\columncolor[gray]{0.9}}cc}
       \toprule
       Smoothing Weight $\beta$ & 0 & 0.2 & 0.4 & 0.8 & 0.95 \\
       \hline
       FID-50K $\downarrow$ & 3.97 & 3.61 & 3.47 & \textbf{3.36} & 3.70 \\
       \bottomrule
    \end{tabular}\vspace{-.4em}
    \label{tab:abl_momentum}
    \begin{minipage}{.96\columnwidth}
       \scriptsize
       \vspace{.8em}
       Action smoothing consistently improves the performance of the larger-step ($T=16$) version of \Ours-MaskGIT-S, with the best performance achieved at $\beta=0.8$.
    \end{minipage}
\end{center}
\vspace{-1.5em}
\end{table}

\paragraph{Ablation Study on the Reward Model Architecture.}
To evaluate the impact of the reward model’s architecture, we compare a convolutional discriminator~\cite{Karras2019AnalyzingAI} and a transformer-based discriminator~\cite{sauer2023stylegan}. As shown in Table~\ref{tab:ablations}(a), both variants significantly outperform the baseline, with the transformer-based design yielding superior results. Based on this observation, we adopt the transformer-based discriminator.

\paragraph{Ablation Study on the Output Activation.}
At the end of our policy network, we introduce an activation function to smoothly constrain outputs within the desired range (detailed in Appendix~\ref{sec:imp_details}). Table~\ref{tab:ablations}(b) compares this approach with simple clamping that directly clips the output range to the desired bounds. Our smooth activation design yields an FID-50K of 4.54, significantly outperforming simple clamping with an FID of 6.29. This demonstrates the effectiveness of smooth constraining activation functions over hard clipping methods.

\paragraph{Ablation Study on the Input Type.}
Table~\ref{tab:ablations}(c) compares using off-the-shelf features from the generative model versus raw intermediate generation results as policy network input. Using features from pre-trained generative models yields substantial improvement with FID-50K of 4.54 versus 6.55 for raw inputs. We attribute this gain to the powerful feature-extraction capabilities of pre-trained generative models, where representations encode semantically meaningful information~\cite{zhao2023unleashing}.

\paragraph{Ablation Study on the Policy Network Architecture.}
Table~\ref{tab:ablations}(d) investigates different architectural choices. Our default conv+MLP configuration achieves FID-50K of 4.54 with 12.0M parameters and 0.016G FLOPs. Replacing convolution with transformer blocks increases cost without improvement. Increasing MLP capacity achieves slight improvement but nearly doubles parameters, while reducing to one linear layer hurts performance with an FID of 5.03.

\paragraph{Ablation Study on the Generation Step Conditioning.}
In our MDP formulation from Section~\ref{sec:ours}, the state space comprises both the intermediate generation result and the current generation step. To quantify the benefit of explicitly incorporating the generation step into the policy network, Table~\ref{tab:ablations}(e) compares the performance with and without step conditioning. Removing generation step conditioning yields significant degradation from FID of 4.54 to 6.13, underscoring its importance. One possible explanation is that effective policy decisions are inherently time-sensitive: since the iterative generative process has a fixed number of generation steps and must terminate after a certain number of iterations, the policy network may need to understand where it stands in this progression to make appropriate interventions.

\section{Conclusion}
We presented \Ours, a unified framework that replaces manually designed scheduling functions in iterative generative models with a learnable policy network, enabling adaptive per-sample customization without requiring expert knowledge. Our key contributions include formulating policy selection as a Markov decision process and introducing an adversarial reward model to prevent overfitting. Comprehensive validation across multiple datasets and paradigms demonstrates significant improvements in generation quality and efficiency. By treating generation policy design as a data-driven optimization problem, \Ours represents an encouraging step toward automated, adaptive generative modeling.

\section*{Acknowledgments}
This work is supported in part by the National Key R\&D Program of China under Grant 2024YFB4708200, the National Natural Science Foundation of China under Grants U24B20173 and the Scientific Research Innovation Capability Support Project for Young Faculty under Grant ZYGXQNJSKYCXNLZCXM-I20.

\bibliographystyle{IEEEtran}
\bibliography{ref}

\appendices
\section*{Appendix}

\section{Experiment Details of Section III}
\label{sec:app_exp_details}
For the experiments in Section~\ref{sec:ours}, we use the MaskGIT-L model on ImageNet 256$\times$256 (see Section~\ref{sec:cin_exp} for details).

In the reward design part of Section~\ref{sec:reward}, we use $T=8$ inference steps.
When implementing FID-based reward design, we find that the FID-based reward model is not able to provide a stable and effective reward signal for the adaptive policy network, leading to divergence:
{
	\begin{center}
		\tablestyle{3pt}{1.1}
		\begin{tabular}{x{80}x{10}|x{40}x{50}}
		\toprule
			& & \multicolumn{2}{c}{FID-50K$\downarrow$} \\
		 Model & $T$ & Adaptive & Non-adaptive \\
		 \hline
		 \Ours-MaskGIT-L & 8 & 55.4 (Fail) & \textbf{2.56} \\
		 \bottomrule
		\end{tabular}
	\end{center}
}
As a result, we adopted a non-adaptive version of the policy network, where all samples share the same generation configuration.
Empirically, the non-adaptive policy network can also be optimized effectively with a low FID.
However, as discussed in Section~\ref{sec:reward}, this numerical superiority does not translate to a practical advantage in terms of sample quality.
The FID reward-based policy network fails to produce images of satisfactory quality.

In Section~\ref{sec:action_stablize}, we use $T=8$ and $T=32$ inference steps for the MaskGIT-L model to study its behavior. For the case with $T=32$, the stabilized policy network adopts an action smoothing factor of $\beta = 0.8$, as suggested by the ablation study in Table~\ref{tab:abl_momentum}.

\section{Implementation Details}
\label{sec:imp_details}
\noindent\textbf{Proximal Policy Optimization (PPO) algorithm.}
\label{sec:ppo}
The Proximal Policy Optimization (PPO) algorithm offers a balance between sample efficiency and ease of implementation.
In this section, we elaborate in more detail on the PPO algorithm adopted in our paper.
First, consider a surrogate objective:
\begin{equation*}
	{L}^{\textnormal{CPI}}(\bm{\phi}) = \mathbb{E}_{t} \left[ \frac{\bm{\pi}_{\bm{\phi}}(\bm{a}_t|\bm{s}_t)}{\bm{\pi}_{\bm{\phi}_{\textnormal{old}}}(\bm{a}_t|\bm{s}_t)} \hat{A}_t \right],
	\label{eq:ppo_cpi}
\end{equation*}
where $\bm{\pi}_{\bm{\phi}}$ and $\bm{\pi}_{\bm{\phi}_{\textnormal{old}}}$ are the policy network before and after the update, respectively.
The advantage estimator $\hat{A}_t$ is computed by:
\begin{equation*}
	\hat{A}_t = -V_{\bm{\phi}}(\bm{s}_t) + R(\bm{s}_T, \bm{a}_T),
\end{equation*}
where $V_{\bm{\phi}}(\bm{s}_t)$ is a learned state-value function.
This objective function effectively maximizes the probability ratio $\rho_t({\bm{\phi}}) = \frac{\bm{\pi}_{\bm{\phi}}(\bm{a}_t \mid \bm{s}_t)}{\bm{\pi}_{{\bm{\phi}}_{\text{old}}}(\bm{a}_t \mid \bm{s}_t)}$ when considering the advantage of taking action $\bm{a}_t$ in state $\bm{s}_t$.
However, directly maximizing ${L}^{\textnormal{CPI}}(\bm{\phi})$ usually leads to an excessively large policy update. Hence, we consider how to modify the objective, to penalize changes to the policy that move $\rho_t({\bm{\phi}})$ away from 1.
This gives rise to the clipped surrogate objective:
\begin{equation*}
	{L}^{\textnormal{CLIP}}(\bm{\phi}) = \mathbb{E}_{t} \Bigg[ \min\left( \rho_t({\bm{\phi}}) \hat{A}_t, \text{clip}\left(\rho_t({\bm{\phi}}), 1-\epsilon, 1+\epsilon\right) \hat{A}_t \right)\Bigg]
	\label{eq:ppo_clip}
\end{equation*}
where $\epsilon$ is a hyper-parameter that controls the range of the probability ratio.
Finally, to learn the state-value function, we additionally add a term for value function estimation error to the objective following~\cite{schulman2017proximal}, which results in the objective in Eq.~\eqref{eq:ppo} in our main paper.

\noindent\textbf{Hyperparameter Details.}
We perform the optimization loop of \Ours{} outlined in Algorithm~\ref{algo:ours} for 1000 iterations. In practice, we perform 5 gradient updates within each iteration of Algorithm~\ref{algo:ours} for both the policy network and the adversarial reward model to facilitate a more stable and efficient optimization process in the minimax game. 
For the optimization of the policy network, we use the Adam optimizer~\cite{kingma2014adam} with a learning rate of $1 \times 10^{-5}$ and betas $(\beta_1, \beta_2) = (0.9, 0.999)$. The clipping parameter in the PPO objective is set to $\epsilon=0.2$, and the value function coefficient is $c=0.5$. The exploration-exploitation hyperparameter $\sigma$ is set to 0.6. The batch size for policy updates is 4096.
For the adversarial reward model, we use the Adam optimizer~\cite{kingma2014adam} with a learning rate of $1 \times 10^{-4}$ and betas $(\beta_1, \beta_2) = (0.5, 0.999)$. The default batch size is 1024.
For experiments on ImageNet 512$\times$512~\cite{russakovsky2015imagenet} and LAION-5B~\cite{schuhmann2022laion}, we reduce the batch size to 512 to fit the memory constraints.
The results on CC3M are based on a publicly available Muse model from github\footnote{https://github.com/baaivision/MUSE-Pytorch}.

\noindent\textbf{Implementation Details of Policy Network.}
We introduce activation functions at the end of the policy network to map the raw, unbounded output to the valid range of each generation policy. Table~\ref{tab:activation_summary} details the specific activation function we used for each generation policy.
Before applying the activation, we also employ the heuristics listed in Table~\ref{tab:hyperparam_summary} as initialization for more effective policy optimization.

\begin{table*}[h]
	\centering
	\caption{Summary of activation functions for different generation policies in \Ours. The policy network produces an unbounded output $x$, which is then passed through an activation function to ensure the resulting policy value lies within its valid range.}
	\label{tab:activation_summary}
	\tablestyle{1pt}{1}
	\begin{tabular}{y{70}y{120}y{90}x{100}}
		\toprule
		Type & Generation Policy & Range & Activation Function \\
		\midrule
		\multirow{4}{*}{MaskGIT}
		& Mask ratio, \(m_t\)  & \(m_t \in [0,1]\)  & $\text{sigmoid}(x)$ \\
		& Sampling temperature, \(\tau_t\)  & \(\tau_t \ge 0\)  & $\text{softplus}(x)$ \\
		& Mask temperature, \(\zeta_t\)  & \(\zeta_t \ge 0\)  & $\text{softplus}(x)$ \\
		& Guidance scale, \(w_t\)  & \(w_t \ge 0\)  & $\text{softplus}(x)$ \\
		\midrule
		\multirow{4}{*}{Autoregressive} 
		& Sampling temperature, \(\tau_t\)  
		& \(\tau_t \ge 0\)  
		& $\text{softplus}(x)$ \\
		& Guidance scale, \(w_t\)  
		& \(w_t \ge 0\)  
		& $\text{softplus}(x)$ \\
		& Top-\(k\) parameter, \(k_t\)  
		& \(k_t \in \mathbb{N}^+\)  
		& $\text{round}(1 + \text{softplus}(x))$ \\
		& Top-\(p\) parameter, \(\rho_t\)  
		& \(\rho_t \in [0,1]\)  
		& $\text{sigmoid}(x)$ \\
		\midrule
		\multirow{2}{*}{Diffusion (ODE)} 
		& Timestep, \(\kappa_t\)  
		& \(\kappa_t \in \{0,\dots,\kappa_{\max}\}\)  
		& $\text{round}(\kappa_{\max} \cdot \text{sigmoid}(x))$ \\
		& Guidance scale, \(w_t\)  
		& \(w_t \ge 0\)  
		& $\text{softplus}(x)$ \\
		\midrule
		\multirow{2}{*}{Rectified Flow} 
		& Timestep, \(\kappa_t\)  
		& \(\kappa_t \in [0,1]\)  
		& $\text{sigmoid}(x)$ \\
		& Guidance scale, \(w_t\)  
		& \(w_t \ge 0\)  
		& $\text{softplus}(x)$ \\
		\bottomrule
	\end{tabular}
\end{table*}

\section{Additional Qualitative Results}
\label{sec:app_additional_qual}
\subsection{Qualitative Results of Inference-time Refinement}
\begin{figure}[htbp]\centering
	\includegraphics[width=\linewidth]{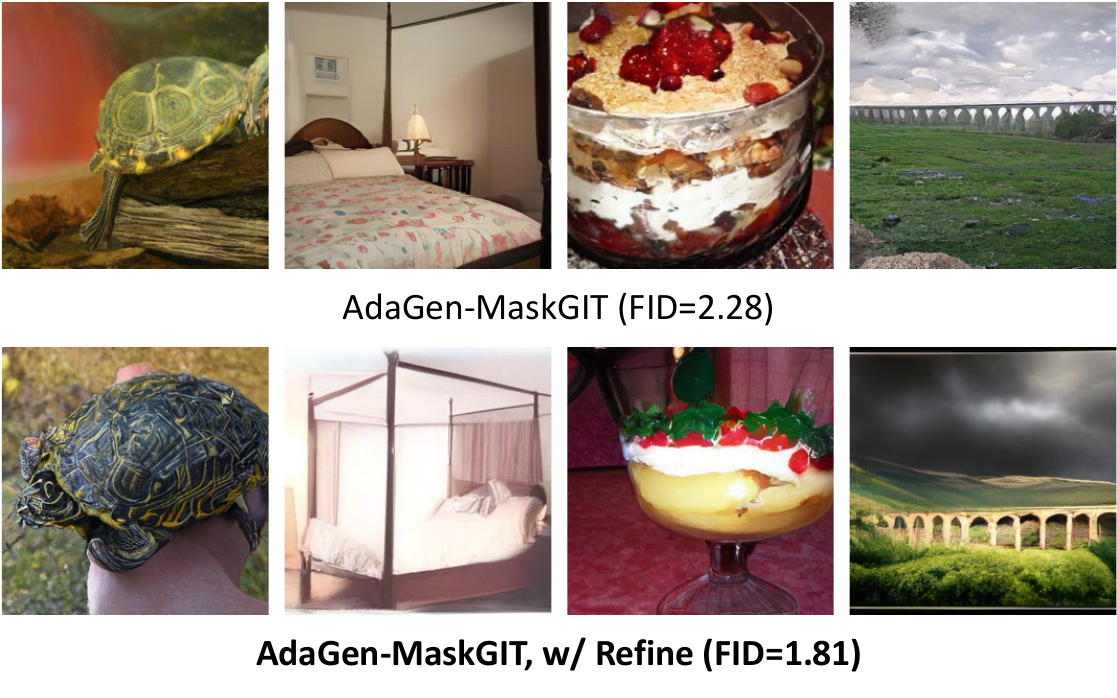}
	\caption{
		{Generated samples without \vs with refinement of \Ours.}
	}
	\label{fig:tts}
\end{figure}

Figure~\ref{fig:tts} presents a qualitative comparison between images generated by AdaGen-MaskGIT ($T=32$) and their refined counterparts obtained through our inference-time refinement strategy. The top row displays the initial outputs of AdaGen-MaskGIT, while the bottom row illustrates the enhanced visual quality achieved after applying our proposed refinement techniques. These visual results corroborate the quantitative improvements reported in Section~\ref{sec:inference_refinement_exp}, demonstrating the capability of our refinement strategy to generate images with higher visual quality.
\subsection{More Generated Samples}
\begin{figure*}[t!]\centering
	\includegraphics[width=\linewidth]{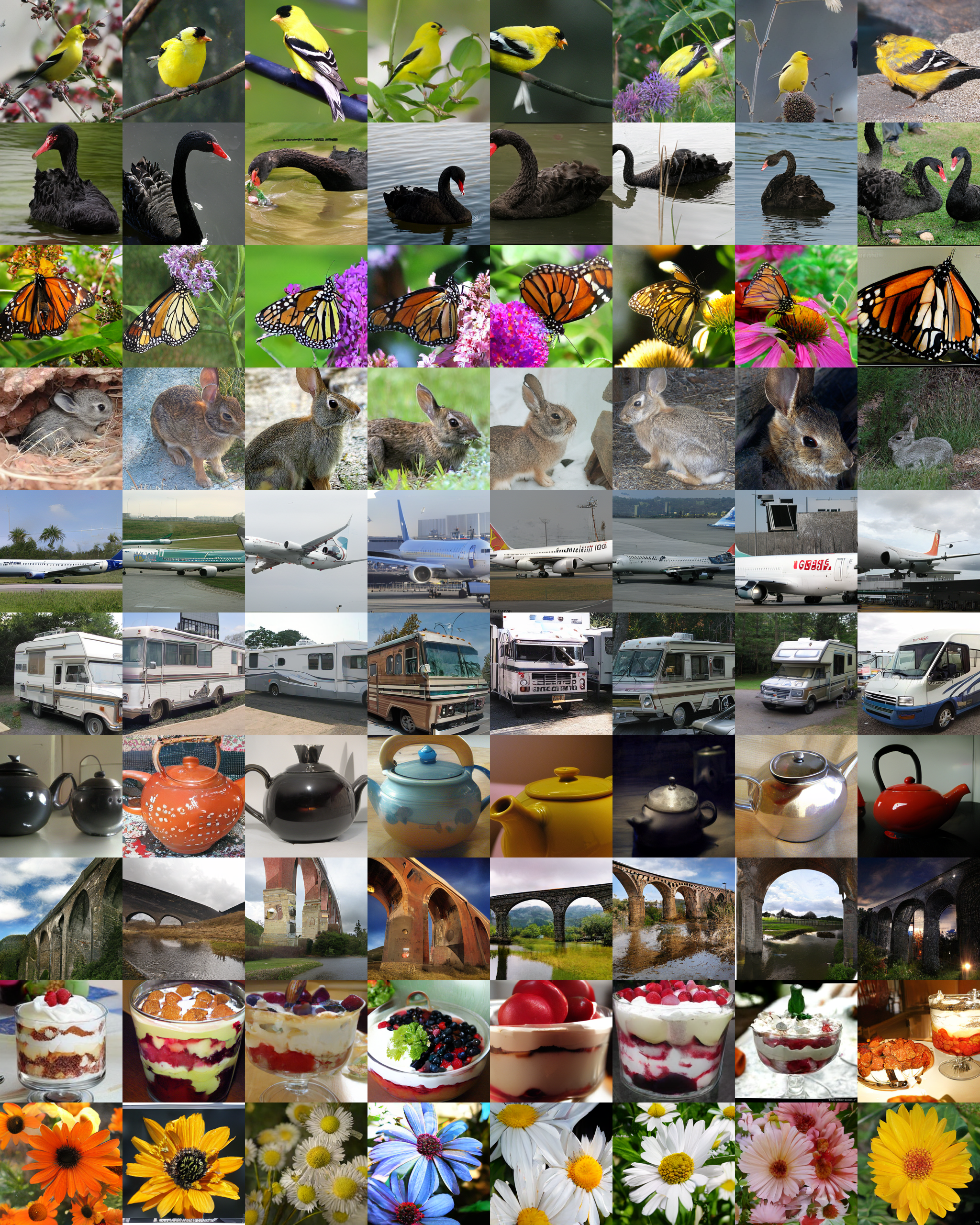}
	\caption{
		{Generated samples of \Ours-VAR-$d$30 ($T=10$) on ImageNet 256$\times$256.}
		\label{fig:more_qual_256}
	}
\end{figure*}

\begin{figure*}[t!]\centering
	\includegraphics[width=\linewidth]{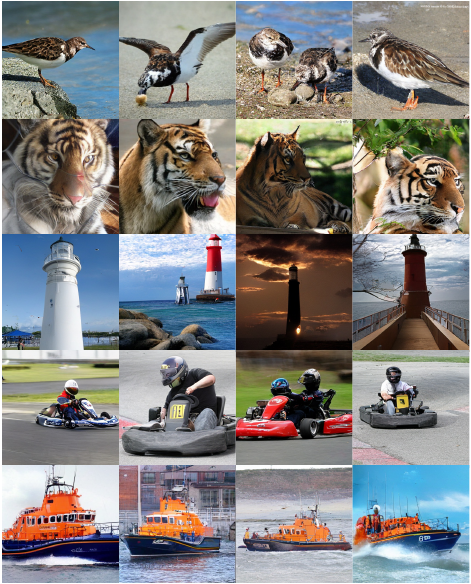}
	\caption{
		{Generated samples of \Ours-MaskGIT-L ($T=32$) on ImageNet 512$\times$512.}
		\label{fig:more_qual_512}
	}
\end{figure*}

\begin{figure*}[t!]\centering
	\includegraphics[width=\linewidth]{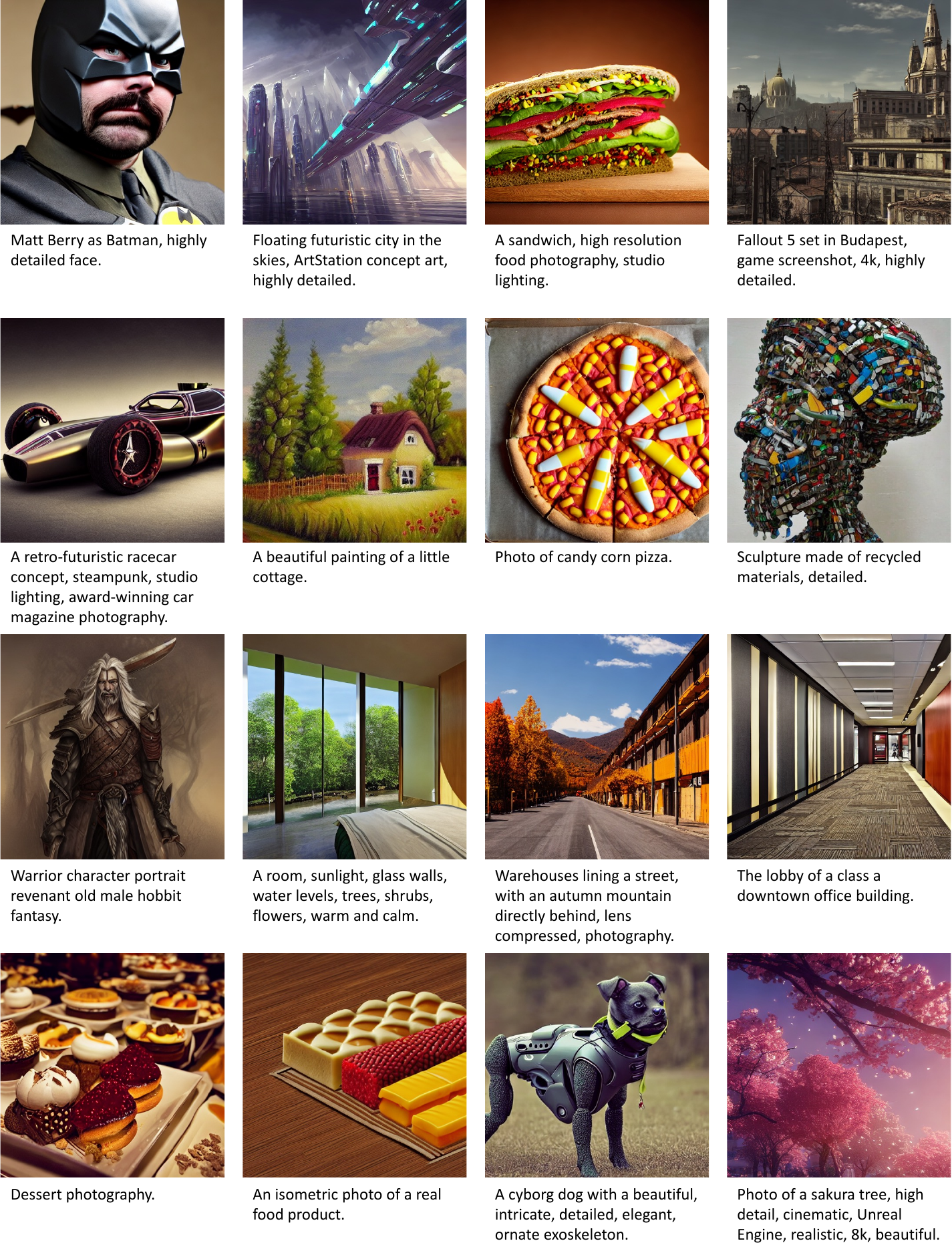}
	\caption{
		{Generated samples of \Ours-Stable-Diffusion ($T=32$) for text-to-image generation.}
		\label{fig:more_qual_t2i}
	}
\end{figure*}

Here we present more generated samples of \Ours on ImageNet 256$\times$256, ImageNet 512$\times$512, and text-to-image tasks.
Figure~\ref{fig:more_qual_256} shows the generated samples of \Ours-VAR-$d$30 ($T=10$) on ImageNet 256$\times$256.
Figure~\ref{fig:more_qual_512} shows the generated samples of \Ours-MaskGIT-L ($T=32$) on ImageNet 512$\times$512.
Figure~\ref{fig:more_qual_t2i} shows the generated samples of \Ours-Stable-Diffusion ($T=32$) for text-to-image generation.

\end{document}